\RequirePackage{fix-cm}
\documentclass[smallextended]{svjour3}       
\smartqed  
\usepackage{graphicx}
\usepackage{float}
\usepackage{comment}
\usepackage[linesnumbered,ruled,vlined]{algorithm2e}
\usepackage{subcaption}
\usepackage{amsmath}
\usepackage{amsmath,amssymb,amsfonts}
\SetKwInput{KwInput}{Input}                
\SetKwInput{KwOutput}{Output}              
\usepackage{authblk}
\usepackage{academicons}
\RequirePackage[colorlinks,citecolor=blue,urlcolor=blue]{hyperref}
\begin{document}

\title{Real-Time Grasping Strategies Using Event Camera
}

\newcommand{\orcidauthorA}{0000-0000-000-000X}
\author{Xiaoqian Huang$^{1}$\href{https://orcid.org/0000-0002-2782-9068}{\includegraphics[scale=1]{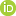}}         \and
        Mohamad Halwani$^{1}$\href{https://orcid.org/0000-0002-8478-2729}{\includegraphics[scale=1]{Figures/orcid.png}}  \and
        Rajkumar Muthusamy$^{1}$\href{https://orcid.org/0000-0002-5372-0154}{\includegraphics[scale=1]{Figures/orcid.png}} \and
        Abdulla Ayyad$^{1,2}$\href{https://orcid.org/0000-0002-3006-2320}{\includegraphics[scale=1]{Figures/orcid.png}} \and
        Dewald Swart$^{3}$ \and
        Lakmal Seneviratne$^{1}$\href{https://orcid.org/0000-0001-6405-8402}{\includegraphics[scale=1]{Figures/orcid.png}} \and
        Dongming Gan$^{4}$\href{https://orcid.org/0000-0001-5327-1902}{\includegraphics[scale=1]{Figures/orcid.png}} \and
        Yahya Zweiri$^{1,5}$\href{https://orcid.org/0000-0003-4331-7254}{\includegraphics[scale=1]{Figures/orcid.png}}
}


\institute{Xiaoqian Huang, Mohamad Halwani, Rajkumar Muthusamy, Abdulla Ayyad, Dewald Swart, Lakmal Seneviratne, Dongming Gan, Yahya Zweiri
              \\
              \email{\{xiaoqian.huang, 100053800,  rajkumar.muthusamy, abdulla.ayyad\}@ku.ac.ae, ASwart@strata.ae, lakmal.seneviratne@ku.ac.ae, dgan@purdue.edu, Y.Zweiri@kingston.ac.uk}       
          \\
          $^{1}$ 
              Khalifa University Center for Autonomous Robotic Systems (KUCARS), Khalifa University, Abu Dhabi, United Arab Emirates \\
             $^{2}$ 
              Aerospace Research and Innovation Center (ARIC), Khalifa University of Science and Technology, Abu Dhabi, UAE.\\
             $^{3}$ 
             Research and Development, Strata Manufacturing PJSC, Al Ain, UAE.\\
            $^{4}$ 
              School of Engineering Technology, Purdue University, West Lafayette, IN 47907, USA \\
            $^{5}$
              Faculty of Science, Engineering and Computing, Kingston University, London SW15 3DW, U.K. \\
              Video link: https://youtu.be/z6MytUIKJAE \\
}

\date{Received: date / Accepted: date}

\maketitle

\begin{abstract}
Robotic vision plays a key role for perceiving the environment in grasping applications. 
However, the conventional framed-based robotic vision, suffering from motion blur and low sampling rate, may not meet the automation needs of evolving industrial requirements. This paper, for the first time, proposes an event-based robotic grasping framework for multiple known and unknown objects in a cluttered scene. Compared with standard frame-based vision, neuromorphic vision has advantages of microsecond-level sampling rate and no motion blur. Building on that, the model-based and model-free approaches are developed for known and unknown objects' grasping respectively. 
For the model-based approach, event-based multi-view approach is used to localize the objects in the scene, and then point cloud processing allows for the clustering and registering of objects. 
Differently, the proposed model-free approach utilizes the developed event-based object segmentation, visual servoing and grasp planning to localize, align to, and grasp the targeting object.  
The proposed approaches are experimentally validated with objects of different sizes, using a UR10 robot with an eye-in-hand neuromorphic camera and a Barrett hand gripper. 
Moreover, the robustness of the two proposed event-based grasping approaches are validated in a low-light environment. This low-light operating ability shows a great advantage over the grasping using the standard frame-based vision. Furthermore, the developed model-free approach demonstrates the advantage of dealing with unknown object without prior knowledge compared to the proposed model-based approach.  
\end{abstract}

\section{Introduction}
\label{intro}

Robots equipped with grippers has become increasingly popular and important for grasping tasks in the industrial field, because they provide the industry with the benefit of cutting manufacturing time while improving throughput. Especially in the $4^{th}$ industrial revolution, the desire for robots that can perform multiple tasks is significant. Assisted by vision, robots are capable to perceive the surrounding environment such as the attributes and locations of the grasping targets. 
The vision-based robotic grasping system can be categorized along various criteria \cite{kleeberger2020survey}. Generally, it can be summarized into analytic and data-driven methods depending on the analysis of the geometric properties of objects \cite{bohg2013data}\cite{sahbani2012overview}. Moreover, according to whether or not building up the object's model, the vision-based grasping can be divided into model-based and model-free approaches \cite{zaidi2017model} \cite{kleeberger2020survey}. Model-based approaches are mostly used for known objects due to the requirement of object's prior knowledge. Model-free methods are more flexible for both known and unknown objects by learning geometric parameters of objects based on vision. Nowadays, with the development of neuromorphic vision in grasping field, the robotic grasping system can be newly categorized into standard vision-based and neuromorphic vision-based approaches along the different perception methods. 
Lots of standard vision-based robotic grasping systems are explored for many applications, such as garbage sorting \cite{zhihong2017vision}, construction \cite{asadi2021automated} and human interaction \cite{ubeda2018vision}.
However, the grasping quality would be affected severely due to the poor perceiving quality, such as the motion blur and poor observing ability in low-light condition. 
On the contrary, neuromorophic vision sensor shows a great potentiality in grasping on account of its high sensitivity, asynchronous property and high sampling rate \cite{gallego2019event}.  
Therefore, we explore the neuromorphic vision-based robotic grasping for multiple objects. 

Robotic vision is one of the most useful sensory functions, but many applications find it difficult to analyze the constant data provided by vision sensors in real time. 
Standard vision sensors continue to sense and save picture data as long as the power is on, resulting in significant power consumption and large data storage. 
Moreover, the low sampling rate affects the efficiency and quality of applications. For example, 
it is proved that the quality of the picture taken by standard camera will be affected by the moving speed of the conveyor belt in production line \cite{zhang2019vision}, due to the motion blur and low sampling rate of the conventional RGB camera. 
In addition, the actuating speed of the electrical gripper is generally over $100ms$ in robotic grasping tasks. Meanwhile, standard cameras commonly have a frame rate of less than 100 per second. Even for the high-speed frame-based camera, the frequency is also generally less than 200 frames per second with a high consumption of both power and storage.
Furthermore, computing the complex algorithm for vision processing algorithm will take additional time to slower the grasping process from the vision resource. So the acceleration of vision acquirement and process will contribute to the grasping efficiency. To improve the reacting speed of vision-based grasping, a faster detecting helps to reserve more time for the gripper’s actuation. For instance, a high sampling rate will assists the robotic system by providing adequate time take actions to prevent slip in a  closed-loop control system. 
Distinct to the conventional frame-based camera, individual events are triggered asynchronously by event camera with a micro second-level sampling rate. 
Therefore, the unique property of event camera 
becomes indispensable to improve the performance for grasping tasks. 

Neuromorphic vision sensors \cite{indiveri2000neuromorphic} differ from conventional vision sensors in that they are inspired by biological systems such as fly eyes, which can sense data in parallel and asynchronously in real time. 
Initially, the neuromorphic vision sensor was known as silicon retina only utilized for computer vision and robotics researches \cite{etienne1996neuromorphic}. Then it becomes known as an event-based camera because it captures per-pixel illumination changes as events \cite{gallego2019event}. Recently, the event-based camera has been utilized in a growing number of applications such as object detection and tracking \cite{mitrokhin2018event}, 3D reconstruction \cite{zhou2018semi}, and simultaneous localization and mapping  \cite{milford2015towards}. 
In contrast to traditional frame-based vision sensors, event-driven neuromorphic vision sensors have low latency, high dynamic range and high temporal resolution. The event camera functions as a neuromorphic vision sensor with the ability to asynchronously measure brightness changes per pixel. It results a stream of events which has microsecond-level time stamp, spatial address, and polarity referring the sign of brightness changes \cite{gallego2019event}. Hence, utilizing events-based segmentation and grasping provides superiorities of  no motion blur, low-light operation, a faster response and higher sampling rate. It introduces new opportunities as well as challenges for neuromorphic vision processing and event-based robotic grasping. 
However, only few works used event camera to address gripping tendencies, such as dynamic force estimation \cite{naeini2019novel} \cite{baghaei2020dynamic}, grasping quality evaluation \cite{huang2020neuromorphic}, and incipient slippages detection and suppression \cite{rigi2018novel} \cite{muthusamy2020neuromorphic}.

The key tasks in robotic vision-based grasping can be summaries as object localization, object pose estimation, grasp generation, and motion planning \cite{du2019vision}. In this work, we assume no obstacles exist between objects and the gripper, so the first three tasks are addressed in the real-time grasping framework. Object localization aims to obtain the target's position, commonly involving object detection and instance segmentation. Using object detection, the object will be classified and located by the bounding box. With the development of deep learning, CNN \cite{chen2019adaptive}, YOLO \cite{redmon2016you} and Faster R-CNN \cite{ren2015faster} are popularly utilized for the object detection. Differently, object instance segmentation is pixel-wise for each individual object. Instance segmentation can be achieved by machine learning based clustering methods, such as KNN \cite{peterson2009k}, K-means \cite{likas2003global}, SVM \cite{cortes1995support} and Mean shift clustering \cite{fukunaga1975estimation}. Local and global masks are another technique generally used in deep learning based instance segmentation, such as YOLACT-the first methods attempting real-time instance segmentation \cite{bolya2019yolact} and SOLO-the segmentation objects by locations \cite{wang2020solo}. However, mostly all the segmentation techniques are applied to standard vision like RGB images and videos. In this work, we develop Multi-object Event-based Mean-Shift instance segmentation for asynchronous event data in model-free approach, and utilize event-based multi-view clustering method in model-based approach. 
As the other part of grasping, the object pose can be estimated by 3D point cloud \cite{zhou2016fast} and image coordinate \cite{hu2020single}. But they are also mostly applied on the standard vision-based grasping, which suffers from motion blur and latency. Grasp generation refers to estimate the grasping pose of gripper, which can be divided into 2D planar grasp and 6DoF Grasp \cite{zhou20176dof} based on standard vision. For event-based grasp generation, the author in \cite{li2020event} constructed Event-Grasping dataset by annotating the best and worst grasp pose using LED's flickering marker. But the frequency is constrained up to $1 KHz$ due to the limitation of LED frequency. Based on the dataset, the grasping angle is learned via deep learning as a "good" or "bad" classification problem. In other words, the gripper is required to adjust pose until the feedback of pose classification achieves "good" class or the stop criteria is reached. Therefore, the grasp pose generation is not efficient since it cannot provide the proper grasp pose directly. 

According to these three tasks of grasping, the neuromorphic eye-in-hand 2D visual servoing approach for single object grasping was developed \cite{muthusamy2021neuromorphic} in our previous work, which adopts an event-based corner detector, a heatmap based corner filter, an event-based gripper alignment strategy. By comparison with the conventional frame-based image-based visual servoing, our previous work shows a superior performance on both time efficiency and computation under different operating speeds and lighting conditions. To improve our prior work for the multiple 3D objects grasping, there are several challenges including the event-based segmentation of multiple objects, the event-based Visual Servoing (EVS) method adoping depth estimation, and the grasp generation according to segmented information represented by spatial-temporal events. 
Therefore, two event-based approaches for multiple objects grasping are developed in this paper, involving the Model-Based Approach (MBA) and Model-Free Approach (MFA). Event-based segmentation, event-based visual servoing adopting depth estimation, and grasping generation using Barrett hand are developed. In addition, we quantitatively evaluate and compare the performance of these two approaches experimentally. 
The primary contributions of the paper are summarized below:
\begin{enumerate}
\item We devise an event-based grasping framework for robotic manipulator with neuromorphic eye-in-hand configuration. In particular, we propose a model-based and model-free approach for grasping objects in a cluttered scene.
\item We study the computational performance of the two event-based grasping approaches and assess their applicability for the real-time and evolving industrial requirements. In particular, we evaluate the grasping framework using a robotic manipulator with a neuromorphic eye-in-hand configuration, a robotic end-effector of Barrett hand, and objects of various sizes and geometries. 
\item We demonstrate the multi-object segmentation, grasping and manipulation for multi-object pick and place applications. In factory automation, the completely event-based strategies can boost production speed. 
\end{enumerate}

Section \ref{sec: overview} introduces the working principle and the data property of event-based camera. Section \ref{sec:model-based} and Section \ref{sec:model-free} elaborate our proposed method for model-based and model-free multi-object grasping using neuromorphic vision, respectively. The validation experiments and results analysis is described in Section \ref{sec: experiment}. Based on the experimental performance, two approaches are discussed and their pros and cons are summarized in Section \ref{sec: discussion}. Then the conclusion and future work are presented in the last section.

\section{Overview of the proposed approaches}
\label{sec: overview}
This section introduces the events data and describes the overall description of Model-Based Approach (MBA) and Model-Free Approach (MFA) using neuromorphic vision 
for robotic grasping, which are elaborated in Section \ref{sec:model-based} and Section \ref{sec:model-free} respectively. 

\begin{figure*}[hbt!]
    \centering
    \centerline{\includegraphics[scale=0.2]{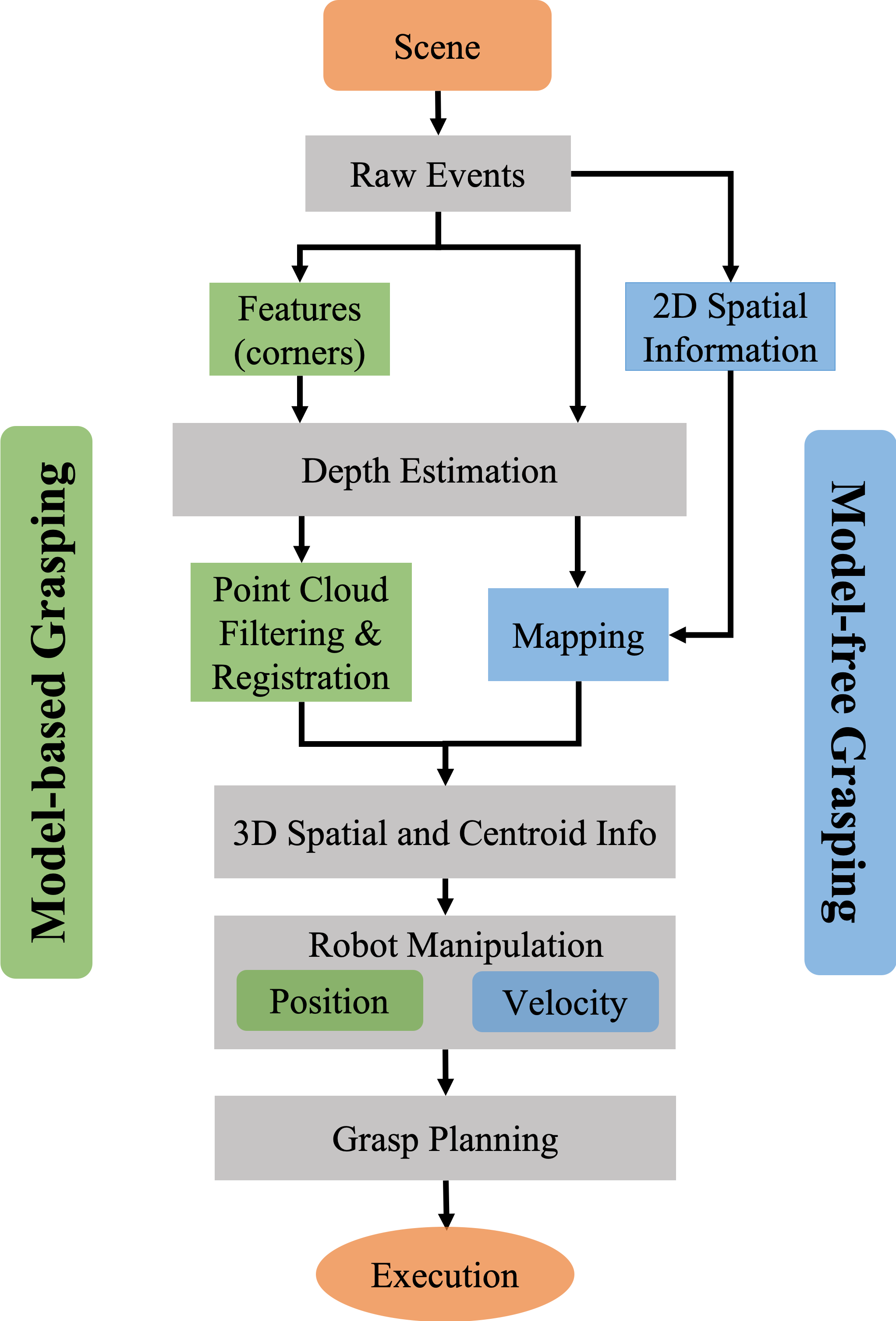}}
    \caption{Model-based grasping (green and gray blocks) and model-free grasping (blue and gray blocks) summary. }
    \label{fig:single pipeline}
\end{figure*}

The pixels of event-based camera can respond to logarithmic brightness $(L = log(I))$ variations independently. Once the perceived logarithmic light intensity change exceeds the threshold $C$, the events will be generated at a pixel $(x, y)$ at time $t$.

\begin{equation}
\begin{split}
\bigtriangleup L(x,y,t) \: = \:
\mid L(x,y,t) - L (x,y,t- \bigtriangleup t) \mid
\\= p *\: C  \:\Bigg\vert \: C>0, p \in \{+1,-1 \}
 \label{eq1}
\end{split}
\end{equation}

where $\Delta t$ is the interval time between the current  and the last event generated at the same pixel, $\Delta L$ represents the illumination change, and $p$ describes positive ($+1$) or negative ($-1$) polarity of events indicating the brightness increase or decrease. The stream of events has a microsecond temporal resolution with an event represented as $e=(x,y,t,p)$ \cite{gallego2019event}.
In this research, DAVIS 346 with a high dynamic range ($140 dB$), low-power consumption ($10 mW$) and $346 \times 260$ resolution will be used. The block diagram of two designed 3D grasping frameworks utilizing event camera is briefly summarized in Fig. \ref{fig:single pipeline}. 

The gray blocks indicate the common processes of the two approaches, and the green and blue blocks represent processes that belong to model-based and model-free approaches, respectively. From the Fig. \ref{fig:single pipeline}, both approaches acquire the 3D spatial and centroid information of individual objects from raw event and depth estimation. Then execute robot manipulation and grasping according to the obtained object's information and grasp planning. The most significant differences contain the segmentation and robotic manipulation methods. The model-based approach segment objects based on the 3D point cloud of features, but the model-free approaches utilized machine learning technique to segment each instance. Moreover, position-based and velocity-based visual servoing are applied for robotic manipulation in the proposed model-based and model-free approach, respectively.

\section{Model-based Grasping Approach}
\label{sec:model-based}
\begin{figure}[h!]
    \centering
    \centerline{\includegraphics[scale=0.2]{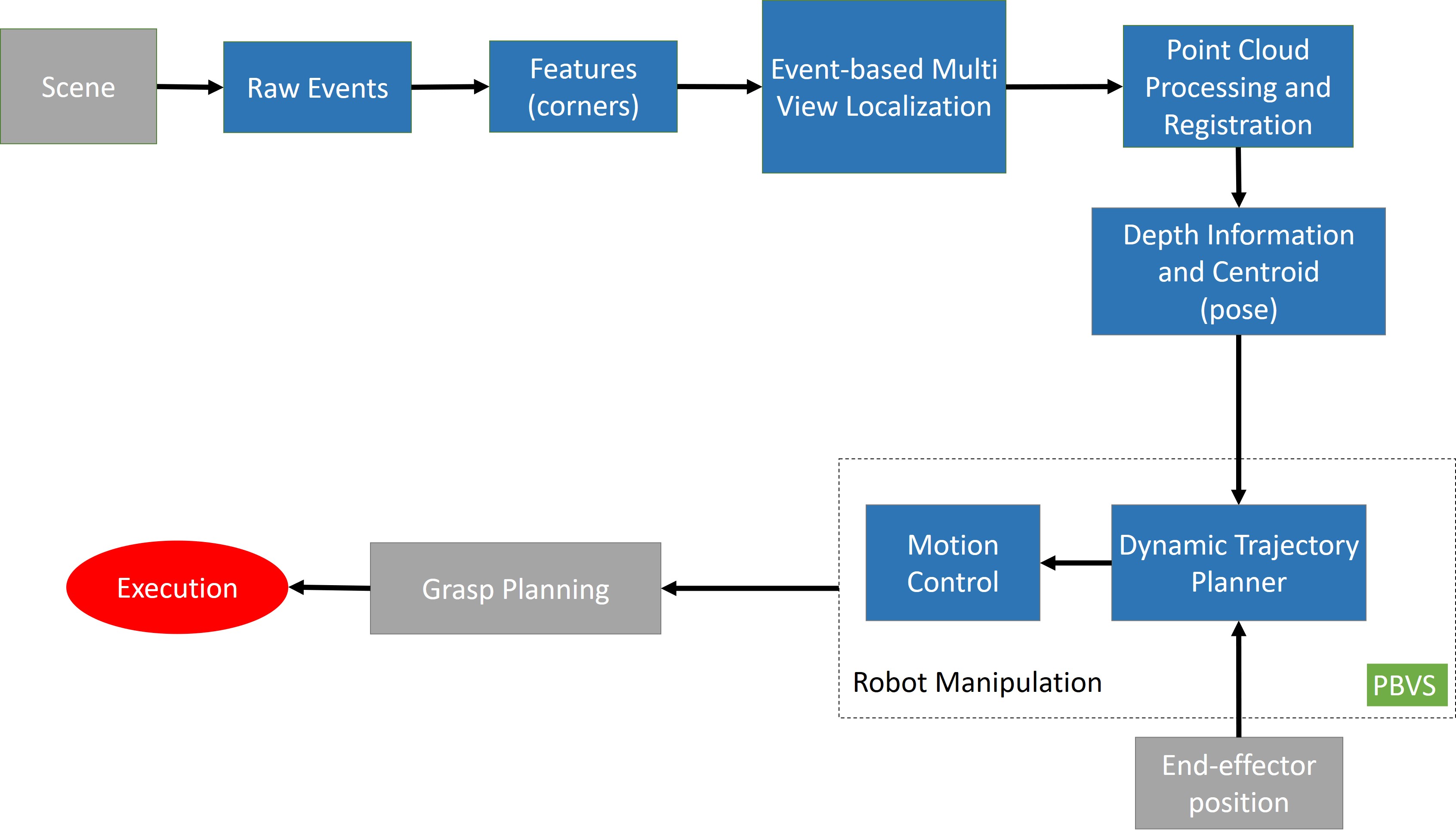}}
    \caption{Model-based multiple-object grasping framework.}
    \label{fig:modelbased-pipeline}
\end{figure}

This section explains an event-based objects grasping approach that uses an inexact-sized model fitting to estimate the pose of the objects to be manipulated. An event camera mounted on robotic manipulator in an eye-in-hand configuration is used. Using pure events from an object in the scene, high-level corner features are extracted using e-Harris corner detector \cite{e-harris}, the use of this detector was justified in our previous work \cite{muthusamy2021neuromorphic} which developed 2D event-based visual servoing approach for single object grasping. As we are using an eye-in-hand setting with a monocular camera, we are missing the depth information of the objects we want to grasp. In the following subsection we will introduce the Neuromorphic Multi-View Localization approach used in this study for the localization of the objects in the environment.  The overall model-based grasping framework is shown in Fig. \ref{fig:modelbased-pipeline}.
\subsection{Neuromorphic Multi-View Localization} \label{EMVS}
\begin{figure}[H]
     \centering
     \centerline{\includegraphics[scale=0.2]{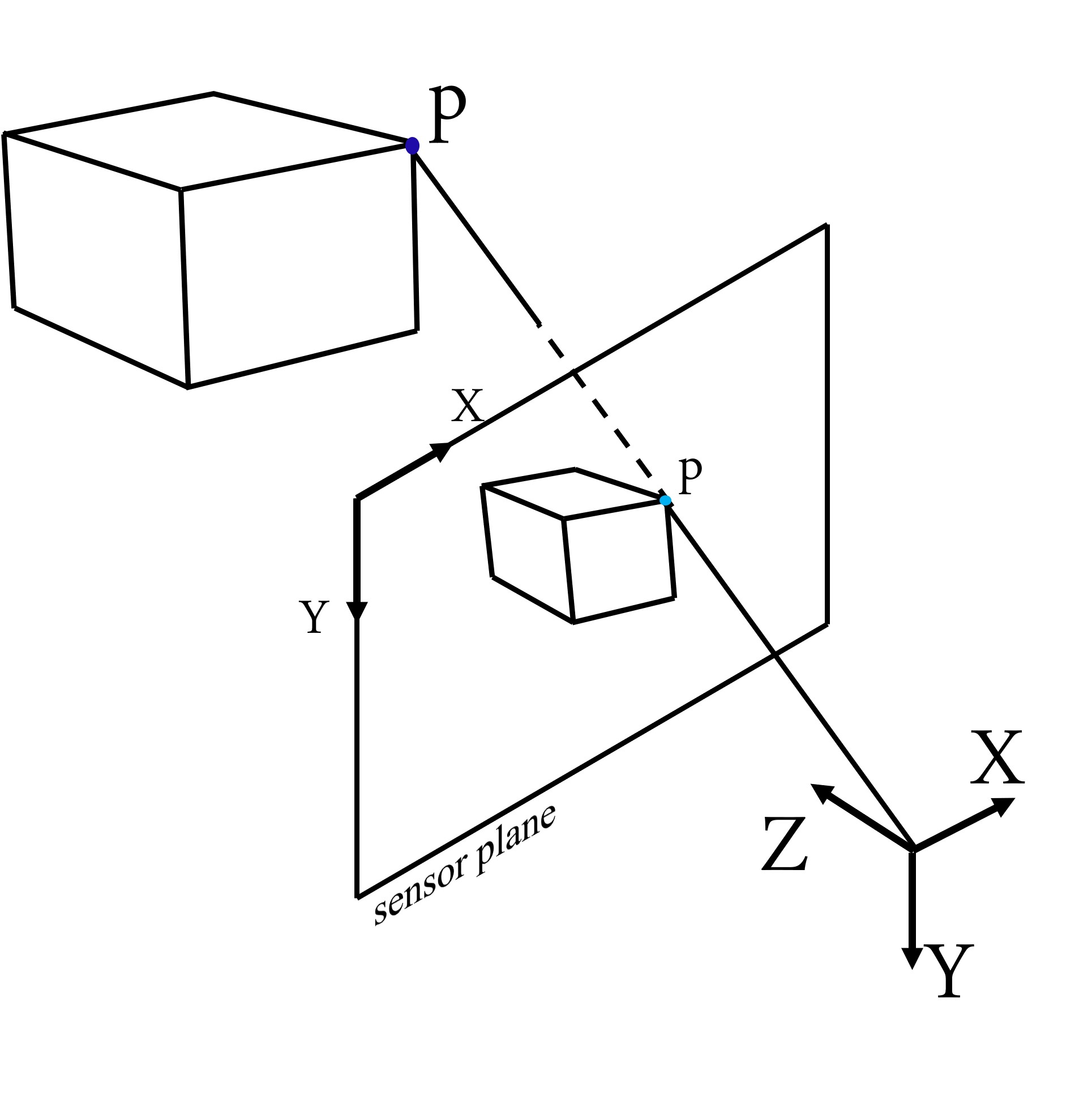}}
     \caption{Relative motion between the camera and the 3D object projects a point (event) to the camera sensor plane}
     \label{fig:pinhole_model}
\end{figure}

Let us consider a moving calibrated event camera observing a rigid static object in the scene. The movement of the camera generates events on the sensor plane of the camera. Event cameras uses the same optica as traditional cameras, so the pingole pojection equation of a 3D point in the environment can be still used. Fig. \ref{fig:pinhole_model} shows the pinhole projection, a 3D point $\textbf{P} = [x,y,z]$ is mapped into a 2D point $\textbf{p} = [u,v]$ on the camera sensor plane, which is expressed as:
\begin{equation}
\bf{z} \begin{bmatrix}u,v,\bf{1}\end{bmatrix}^T =\quad  \bf{K}   
    \begin{bmatrix}
    \bf{R} & \bf{t}  \end{bmatrix} 
    \begin{bmatrix} \bf{x}, \bf{y}, \bf{z}, \bf{1} \end{bmatrix}^T
 \label{eq1}
\end{equation}
where K is a $3 \times 3$ camera's intrinsic parameters and R and t are the relative pose between the camera and the object in the environment.
The most common approach that tackles objects localization using event-cameras is by using two or more event cameras with known fixed attachment between them, and sharing a common clock. This method requires to solve for the events correspondence among the two cameras and then localize the point feature in the environment. In our work we used an approach that considers only a single camera with a known camera trajectory. Event-Based Multi-View Stereo (EMVS) introduced in \cite{rebecq2016emvs}, was used to estimate the exact pose of the object but utilizing only the high-level corner features and the information of the known trajectory of the event camera.  The benefit of this approach is that it considers the asynchronous and sparse nature of the events generated by the event camera to warp them as rays through a discretized volume of interest called the Disparity Space Image (DSI). This method works directly in 3D space, so it is not requried to recover the 3D location by associating individual event to a particular 3D point. The event stream \(e_{k}\) generated by the event camera is the input point features that are warpped into the DSI as rays according to the viewpoint of the event camera at time \(t_k\). Then, the number of rays passing through each voxel is counted and a 3D point is considered present or not in each voxel by voxel voting. The approach back-projects the events from their corresponding camera view to a reference camera view to find their exact projection location in depth plane $Z_i$. An event is back projected via two steps: first, it is mapped to the first depth plane $Z_0$ using the homography $H_{Z_0}$ and then to its correspondent depth plane $Z_i$ using the homography $H_{Z_i} H^{-1}_{Z_0}$. where:

\begin{equation}
    \label{eq_homography}
    H_{Z_i} \sim R + \frac{1}{Z_i}t\bf{e} ^T_3
\end{equation}
and 
\begin{equation}
    \label{eq_homography}
   \bf{e_3} = (0,0,1)
\end{equation}

The back projected viewing rays passing through the DSI are counted for each voxel. First the DSI is defined by the camera pixels and a pre-defined number of depth planes $N_z$, therefore, it has a size of $width \times height \times N_z$. The amount of viewing rays  that intersect each voxel are stored as a score in the DSI: 
\begin{equation}
    f(\boldsymbol{X}): V \subset \mathbb{R}^3
\end{equation}
where $\boldsymbol{X} = (X,Y,Z) ^ T$ represents voxel center. 
Finally, EMVS algorithm produces a semi-dense depth map by thresholding a confidence map $c(x,y)$, which represents the location and magnitude of the optimal score with the maximum value as follows:
\begin{equation}
   f(X(x), Y(y), Z^*) =: c(x,y)
\end{equation}
The depth map can be then converted to a point cloud.

\subsection{Point Cloud Processing}
The model-based multi-objects grasping is achieved by executing the following: point cloud downsampling, object clustering, model registration, and robot manipulation with grasp planning, which are explained in the following subsections respectively.

\subsubsection{Point Cloud Downsampling} \label{subsec: Downsampling}
Post processing of the point cloud is performed to remove the outliers in the point cloud. Using a space partitioning data structure for organizing the points in a k-dimensional space in a tree structure (k-d tree), we used a nearest neighbour search to remove points that have a distance higher than a threshold. This helps removing isolated points which are most likely outliers.

\subsubsection{Object Clustering} \label{subsec: clustering} 
Grasping multiple objects requires finding and segmenting the objects in the scene to individual object point clusters. Since our point clouds contain only objects corner points (i.e., maximum 8 points per object), we can use simple data clustering approaches without worrying about the execution speed of the clustering algorithm. We applied an euclidean cluster extraction method, implemented using point cloud library (PCL) \cite{PCL}. This method divides an unorganized point cloud model into smaller parts to reduce the processing time. Same as the method used in Section \ref{subsec: Downsampling}, a tree data structure is used to subdivide the points and make use of the nearest neighbours to search for points that are within a sphere of radius equal to a threshold and adds them to a point cloud cluster $\{C_i\}_{i=1}^N$, where $N$ is the number of objects detected.

\subsubsection{Model Registration} \label{subsubsec:modelREG}

Because of camera visual constraints and objects geometrical constraints, not all object corners can be viewed considering the linear motion of the eye-in-hand, thus, this issue has to be solved to provide an exact model of the object to be grasped. Object registration aligns two point clouds to find the relative pose between the two point clouds in a global coordinate frame. Iterative Closest Point (ICP) offers a good solution to solve for the un-seen corners and performing model registration, but original ICP needs an exact model of the targeted object to find the transformation between the target model point cloud $C_i$ and the source model point cloud $P$, thus, it does not handle the case of models with different scales. In addition, to increase the chance of good convergence and successful alignment using ICP, an initial rough estimate of the alignment is required  to avoid converging in a local minima \cite{ICP}. However, finding a rough estimate requires finding feature descriptors to determine point-to-point correspondences, yet its a challenging problem since we operate on a small sized point clouds, and common feature based descriptors (i.e., spin image, PFH, DH, etc.) were designed for dense point clouds. 

In our paper, we used an inexact model $P$ (i.e., 8 corners relevant to a cube of length 1 m) to generalize our registration algorithm. According to \cite{sankaranarayanan2007fast}, the registration problem is divided into 4 steps:
\begin{enumerate}
\item Selection: select input point cloud.
\item Matching: estimate correspondences.
\item Rejection: filter to remove outliers.
\item Alignment: find optimal transformation by minimizing an error metric.
\end{enumerate}

As discussed in Section \ref{subsec: clustering} and Section \ref{subsec: Downsampling}, the input point cloud is the clustered point cloud $C_i$ after removing the isolated points. 
We used Singular Value Decomposition (SVD), to find the optimal transformation parameters (rotation R, translation t and scaling c) between the set of points \(X = {x_1, x_2, \cdots, x_n}\) in cluster $C_i$ and the set of points \(Y = {y_1, y_2, \cdots, y_n}\) from the source model after solving the correspondence, where each set of points of m-dimensional space (i.e., 3D in our case). The mean squared error of these two point sets is 

\begin{equation}
    \label{eq:32}
    e^2(R,t,c) = \frac{1}{n} \sum_{i=1}^{n} \left\|{y_i - (cRx_i + t)}\right\|^2
\end{equation} 
where
\begin{equation}
    \label{eq:34}
    \mu_x = \frac{1}{n} \sum_{i=1}^{n} x_i
\end{equation} 

\begin{equation}
    \label{eq:35}
    \mu_y = \frac{1}{n} \sum_{i=1}^{n} y_i
\end{equation} 

\begin{equation}
    \label{eq:36}
    \sigma_x^2 = \frac{1}{n} \sum_{i=1}^{n} \left\|{x_i - \mu_x}\right\|^2
\end{equation} 

\begin{equation}
    \label{eq:37}
    \sigma_y^2 = \frac{1}{n} \sum_{i=1}^{n} \left\|{y_i - \mu_y}\right\|^2
\end{equation} 

\begin{equation}
    \label{eq:38}
    \sum_{xy} = \frac{1}{n} \sum_{i=1}^{n} (y_i - \mu_y)(x_i - \mu_x)^T
\end{equation} 
and let the SVD of equation \ref{eq:38} be $UDV^T$ where D is a diagonal matrix of size m and 
\begin{equation}
    \label{eq:39}
    S =  \begin{cases}
      I & \quad \text{if } det(\sum_{xy}) \geq 0 \\
      diag(1, 1, \cdots, 1, -1) & \quad  \text{if } det(\sum_{xy}) < 0.
    \end{cases}
\end{equation} 
where $\sum_{xy}$ is a covariance matrix of $X$ and $Y$, $\mu_x$ and $\mu_y$ are mean vectors of \(X\) and \(Y\), and  $\sigma_y^2$ and $\sigma_y^2$ are variances around the mean vectors of $X$ and $Y$, respectively.
Hence, the optimal transformation variables can be computed as follows:
\begin{equation}
    \label{eq:40}
    R = USV^T
\end{equation}

\begin{equation}
    \label{eq:41}
    t = \mu_y - cR\mu_x
\end{equation}

\begin{equation}
    \label{eq:42}
    c = \frac{1}{\sigma_x^2}tr(DS)
\end{equation}

For mathematical proof of the equations you can review \cite{SVD}.

\subsection{Robot Manipulation and Grasp Planning}

The robot manipulation controller for the model-based object grasping is controlled by a Position Based Robot Controller (PBVS). The PBVS stage guides the end-effector towards the object features using the 6DoF pose estimate from the multi-view detection and the model registration stage explained in \ref{subsubsec:modelREG}. PBVS considers a known intial and final poses of the robot's end effector. The final pose can be pre-defined as in the detection step, or it is found  by the EMVS and the point cloud processing stages, which would represent the object centroid and orientation angle. The desired joint angles vector of the robotic arm $\hat{\theta}$ is found using an inverse kinematic approach of the open-source Kinematic and Dynamics Library. Afterwards, we compute a trajectory for the joint angles on the Open Motion Planning Library, and a PID controller regulates each joint and tracks the joint angles.  

\begin{figure}[!h]
\begin{algorithm}[H]
\DontPrintSemicolon

  \KwInput{Events stream: position $(x_i, y_i)$, timestamp  $t_i$}
  \KwOutput{Objects centroid $(X,Y,Z)$, Object Orientation Angle}
Set starting point of the gripper $P_0$\\
Scan the scene and extract objects corners\\ 
Perform the event-based multi-view localization\\
Perform point cloud downsampling\\
Perform object Euclidean clustering\\
    \For{ each $cluster$}{ 
Perform model registration\\
Extract object centroid\\
Extract object orientation\\ 
Navigate to object and orient the gripper\\
Perform Grasp\\
    }

\caption{Model-based Grasping Framework}
\end{algorithm}
\label{alg:model-based}
\end{figure}

\section{Model-Free Event-based Multiple Objects Grasping}
\label{sec:model-free}

Distinct from the model-based approach, the model-free approach acquires object information directly from the asynchronous event data. For 3D object detection and grasping, the depth information can be obtained by EMVS, and mapped into image coordinates for further visual servoing. Building on that, the velocity control will be utilized to manipulate the robot arm, ensuring the end-effector/gripper is aligned with the centroid. While the grasp hypothesis is generated for the gripper according to the event stream, then grasping operation will be executed to enclasp the aiming object. Therefore, the designed event based multi-object grasping consists of three parts: segmentation, visual servoing, and grasping plan, which are explained in the following subsections respectively. 


\subsection{Segmentation}

\begin{figure}[!ht]
    \centering
    \centerline{\includegraphics[scale=0.6]{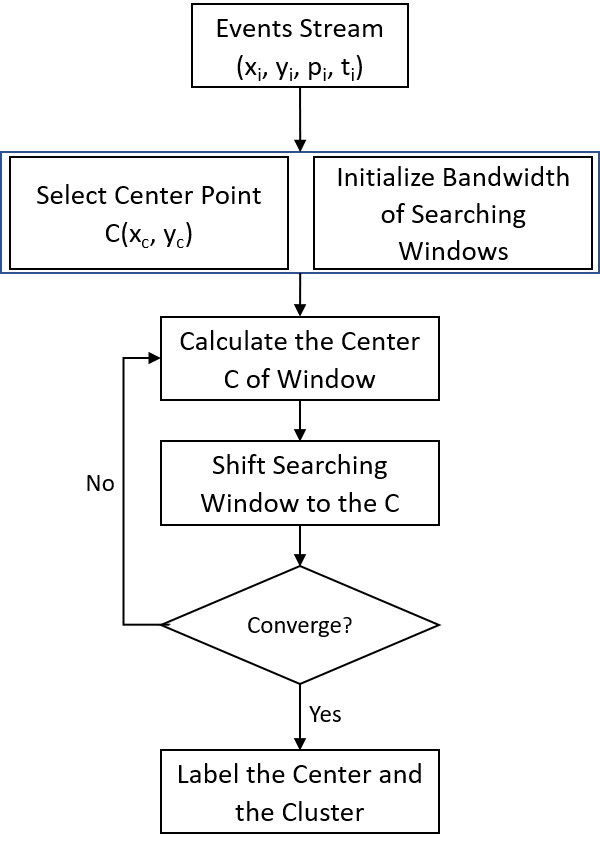}}
    \caption{Event-based mean shift clustering principle}
    \label{fig:MS-flow-chart}
\end{figure}

In this work, the Mean Shift (MS) algorithm is employed for object segmentation due to its non-parametric character, which is developed based on the assumption that different clusters of data are of different probability density distributions. The working principle is that by computing the shifting vectors of one point and all neighbouring points in some range, the mean shift vector can be obtained including shifting magnitude and orientation. Then repeating calculating this mean shift vector until it converges. The main idea of object segmentation based on events data is visualized in Fig. \ref{fig:MS-flow-chart}. 

The probability density distributions can be expressed as the Probability Density Function (PDF) shown in Equation \ref{eq:PDF}: 
\begin{equation}
P_{x}=\frac{1}{nh^2}\sum_{i=1}^{n} K \frac{(x-x_i) }{h^2}
\label{eq:PDF}
\end{equation}
where $K$ is the kernel function applied to each data point, $n$ denotes the number of data points, and $h$ presents the bandwidth parameter which means the radius of the kernel. 

For dealing with the non-linear datasets, the data is usually reflected into high dimension space by using kernel function $K(x)$. The most used kernel- Gaussian kernel is expressed in Equation \ref{eq:gaussain_kernel}.

\begin{equation}
K_G(x)=e^{ -\frac{x^2}{2\sigma^2} }
\label{eq:gaussain_kernel}
\end{equation}

where $\sigma$ is the bandwidth of the window. In mean shift procedure, each point has its own weight and bandwidth which can be an isotropic and diagonal matrix. To simplify the expression, the case that all points have the same and scalar bandwidth $\sigma$ and the same weight $\frac{1}{nh^2}$ is considered most practically. In addition, the Gaussian kernel is mostly used in PDF because the bandwidth $\sigma$ will be the only parameter required for MS clustering. 

Suppose $x$ is a point to be shifted and $N(x)$ are the sets of points near the point $x$. $D(x,x_i)$ is the distance from the point $x$ to the point $x_i$, then the new position $x^\prime$ shifted from $x$ is calculated as Equation \ref{eq:MS:shifted_position}. 

\begin{equation}
x^\prime=\frac{\sum_{x_i\subset N(x)}k(D(x,x_i)^2)x_i}{\sum_{x_i\subset N(x)}k(D(x,x_i)^2)}
\label{eq:MS:shifted_position}
\end{equation}

The new position will keep updating until the MS vector is converged or the maximum iteration step is reached. The MS vector is represented as:
\begin{equation}
V_x=\frac{\sum_{x_i\subset N(x)}k(D(x,x_i)^2)x_i}{\sum_{x_i\subset N(x)}k(D(x,x_i)^2)}-x
\label{eq:MS:vector}
\end{equation}

The standard MS algorithm only considers the spatial information of data points. For dealing with the asynchronous and sparse evens data, both spatial and temporal information is used for multiple objects in this research as Multi-object Event-based Mean Shift (MEMS). 
The main idea of instance segmentation based on events data is shown in Fig. \ref{fig:MS-flow-chart}. By repeatedly updating a given point with the mean of the shifting vectors with respect to all its neighbouring points within a specified range, the process will eventually converge to the distribution mode of the cluster to which the starting point belongs.
Currently, MEMS algorithm is applied on 2D spatio-temporal events stream which are represented as $(x_i,y_i,t_i)$, where $(x_i,y_i)$ and $t_i$ are spatial and temporal information respectively. The bandwidth of the searching window is initialized before processing. The spatial center point $c(x_i,y_i)$ will be randomly selected after obtaining events data. Then iterating the procedure of mean computing and shifting until it is converged.

Utilizing both spatial and temporal information, the event-based PDF and Gaussian kernel are expressed as Equation \ref{eq:pdf_pt}, Equation \ref{eq:gaussain}:
\begin{equation}
P_{\boldsymbol x,t}=\frac{1}{n}\sum_{i=1}^{n} K ( \left[\boldsymbol x,t\right]-\left[\boldsymbol x_i,t_i\right])
\label{eq:pdf_pt}
\end{equation} 

\begin{equation}
K(\left[\boldsymbol x,t\right]-\left[\boldsymbol x_i,t_i\right])=ck(\mid\mid\frac{\left[\boldsymbol x,t\right]-\left[\boldsymbol x_i,t_i\right]}{\sigma}\mid\mid)^2=ce^{-\frac{(\boldsymbol x-x_i)^2+(t-t_i)^2}{2\sigma^2} }
\label{eq:gaussain}
\end{equation}

where $\boldsymbol x_i$ is a 2D vector representing the spatial coordinates, $t_i$ is the time stamp of \textbf{$x_i$}, and $c$ is the coefficient of Gaussian kernel which is equal to $\frac{1}{\sqrt{2\pi}\sigma}$. 
Underlying density $P_{\boldsymbol x,t}$ is to find the modes of this density. The modes are located among the zeros of the gradient $\nabla$ $P_{\boldsymbol x,t}=0$ and the mean shift procedure is an elegant way to locate these zeros without estimating the density. 
The $\nabla$ $P_{\boldsymbol x,t}$ is obtained as:
\begin{equation}
\nabla P_{\boldsymbol x,t}=\frac{1}{n}\sum_{i=1}^{n} \nabla K(\left[\boldsymbol x,t\right]-\left[\boldsymbol x_i,t_i\right])
\label{eq:MS:delta_p1}
\end{equation}
By substituting Equation \ref{eq:gaussain} into Equation \ref{eq:MS:delta_p1}, $\nabla P_{\boldsymbol x,t}$ is obtained as:
\begin{equation}
\nabla P_{\boldsymbol x,t}=\frac{c}{n}\sum_{i=1}^{n}g_i \left[\frac{\sum_{i=1}^{n}\left[\boldsymbol x_i,t\right] g_i}{\sum_{i=1}^{n}g_i}-\left[\boldsymbol x,t\right] \right]
,
g=-k^\prime
\label{eq:MS:delta_p2}
\end{equation}
Then the MS vector can be expressed as Equation \ref{eq:MS_vector}, which is important for target localization and gripper manipulation.
\begin{equation}
\boldsymbol m(\left[\boldsymbol x,t\right])= \left[\frac{\sum_{i=1}^{n}\left[\boldsymbol x_i,t\right] g(\frac{\mid\mid\left[\boldsymbol x,t\right]-\left[\boldsymbol x_i,t_i\right]\mid\mid^2}{h})}{\sum_{i=1}^{n}g(\frac{\mid\mid\left[\boldsymbol x,t\right]-\left[\boldsymbol x_i,t_i\right]\mid\mid^2}{h})}-\left[\boldsymbol x,t\right] \right]
\label{eq:MS_vector}
\end{equation}Running the MEMS on the event data obtained by the neuromorphic sensor, the execution time is $12.861ms$ presenting a dramatic improvement of time efficiency compared to the mean shift algorithm for standard vision (754.977 $ms$). 
To further accelerate the segmenting speed of MEMS, we applied two strategies: soft speedup term and downsampling data illustrated in Section \ref{subsec:strategy1} and \ref{subsec:strategy2}, respectively. For robotic grasping tasks, both efficiency and accuracy are key aspects for evaluating the performance of MEMS. Therefore, the metric E-score is designed as the following equations: 

\begin{equation}
    E-score=\lambda_1 \cdot Ere + \lambda_2 \cdot Fre
    \label{eq:e-score}
\end{equation}
where $\lambda_1$ and $\lambda_2$ are factors indicating the significance of efficiency and accuracy considered in the task, and the sum of $\lambda_1$ and $\lambda_2$ is constrained as 1 that $\lambda_1 + \lambda_2 = 1$. Compared with the baseline when $\alpha=0$ in strategy 1 or $\beta=1$ in strategy 2, $Ere$ (Eq. \ref{eq:Ere}) and $Fre$ (Eq. \ref{eq:Fre}) represent the relative error of processing time per event and F1 score, respectively. 
\begin{equation}
    Ere=-\frac{T_e(\alpha or \beta)-T_e(\alpha=0 or  \beta=1)}{T_e(\alpha=0 or \beta=1)}\cdot 100
    \label{eq:Ere}
\end{equation}
\begin{equation}
    Fre=-\frac{F1(\alpha or \beta)-F1(\alpha=0 or \beta=1)}{F1(\alpha=0 or \beta=1)}\cdot 100
    \label{eq:Fre}
\end{equation}

where $T_e$ and $F1$ are the processing time per event and F1 score correspondingly. Time efficiency is the main concern in this work, so the core contribution of MEMS is to accelerate the standard mean shift. Therefore, $\lambda_1$ and $\lambda_2$ are set as 0.6 and 0.4 respectively to assess the overall performance.

\subsubsection{Strategy 1}
\label{subsec:strategy1}
\begin{figure*}[!ht]
    \centering
    \hspace{-1.4cm}
    \begin{subfigure}[t]{0.58\textwidth}
    \centerline{\includegraphics[height=1.9in]{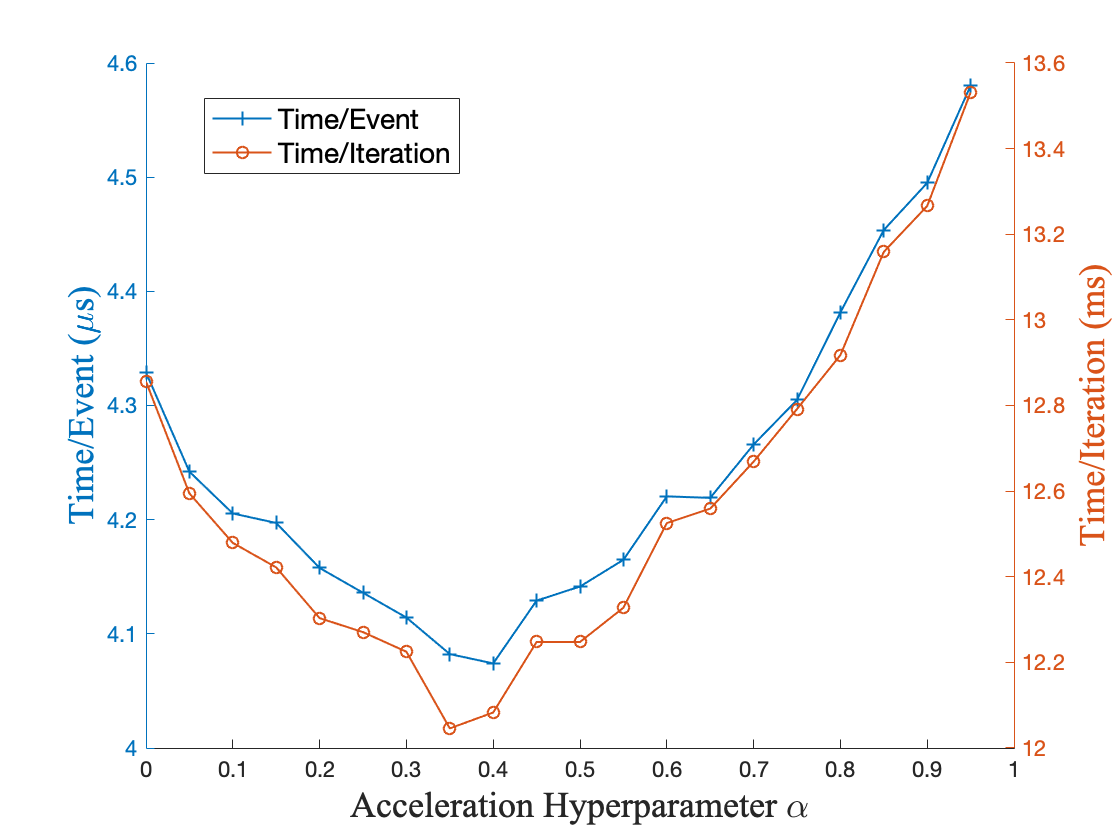}}
    \caption{Processing time}
    \label{fig:alpha-time}
    \end{subfigure}
    ~
    \begin{subfigure}[t]{0.4\textwidth}
    \centerline{\includegraphics[height=1.9in]{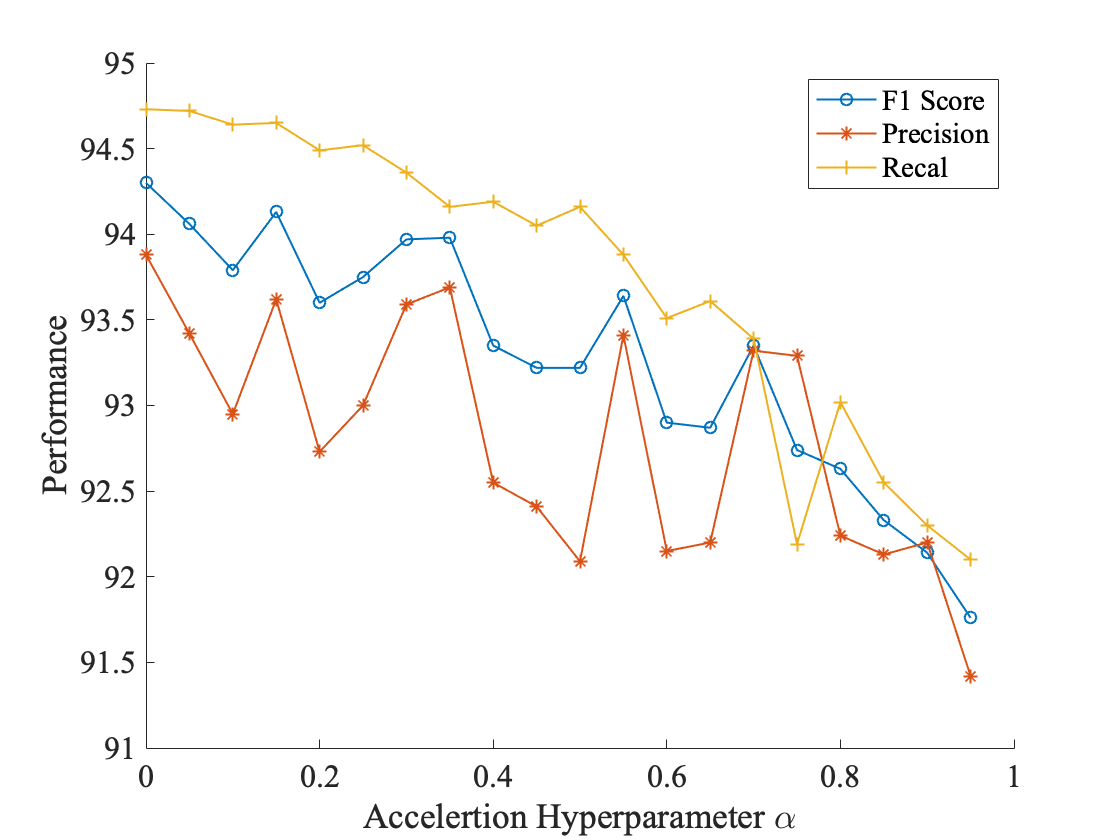}}
    \caption{Performance}
    \label{fig:alpha-performance}
    \end{subfigure}
    \caption{Processing time and performance with varying acceleration hyperparameters $\alpha$}
    \label{fig:alpha}
\end{figure*}

In iterations of MEMS, each event will be shifted along the shifting vector with the magnitude. The speedup term is then added to calculate the final new positions as expressed in Equation \ref{eq:MS_new}. 
\begin{equation}
\boldsymbol x^\prime= \left[\frac{\sum_{i=1}^{n}\left[\boldsymbol x_i,t\right] g(\frac{\mid\mid\left[\boldsymbol x,t\right]-\left[\boldsymbol x_i,t_i\right]\mid\mid^2}{h})}{\sum_{i=1}^{n}g(\frac{\mid\mid\left[\boldsymbol x,t\right]-\left[\boldsymbol x_i,t_i\right]\mid\mid^2}{h})}+\alpha \cdot \boldsymbol{m} \right]
\label{eq:MS_new}
\end{equation}
where $\alpha$ is the acceleration coefficient that controls how much extra distance the shifted points move along the shift vector. While hyperparameter $\alpha$ is set properly, the procedure will be accelerated compared with the standard MS. We tested the mean shift speed for individual events and iteration using different $\alpha$ within 1 as shown in Fig. \ref{fig:alpha}(a).


As shown in Fig. \ref{fig:alpha}(a), the processing time occupied by each event and iteration demonstrates a similar pattern, that declines when $\alpha \leq 0.35$ and increases while $\alpha >0.35$, and almost the smallest deviations are demonstrated when $\alpha = 0.35$. In other words, the efficiency of MEMS in event and iteration levels will be improved within some range of $\alpha$. However, the MEMS will diverge when $\alpha$ reaches 1. Besides, the clustering accuracy of MEMS is assessed by precision, recall and F1 score as recorded in Fig. \ref{fig:alpha}(b). 
Although three metrics demonstrate slightly better when $\alpha=0$, the average differences are only around $1\%$.

\begin{figure}[!ht]
    \centering
    \centerline{\includegraphics[scale=0.19]{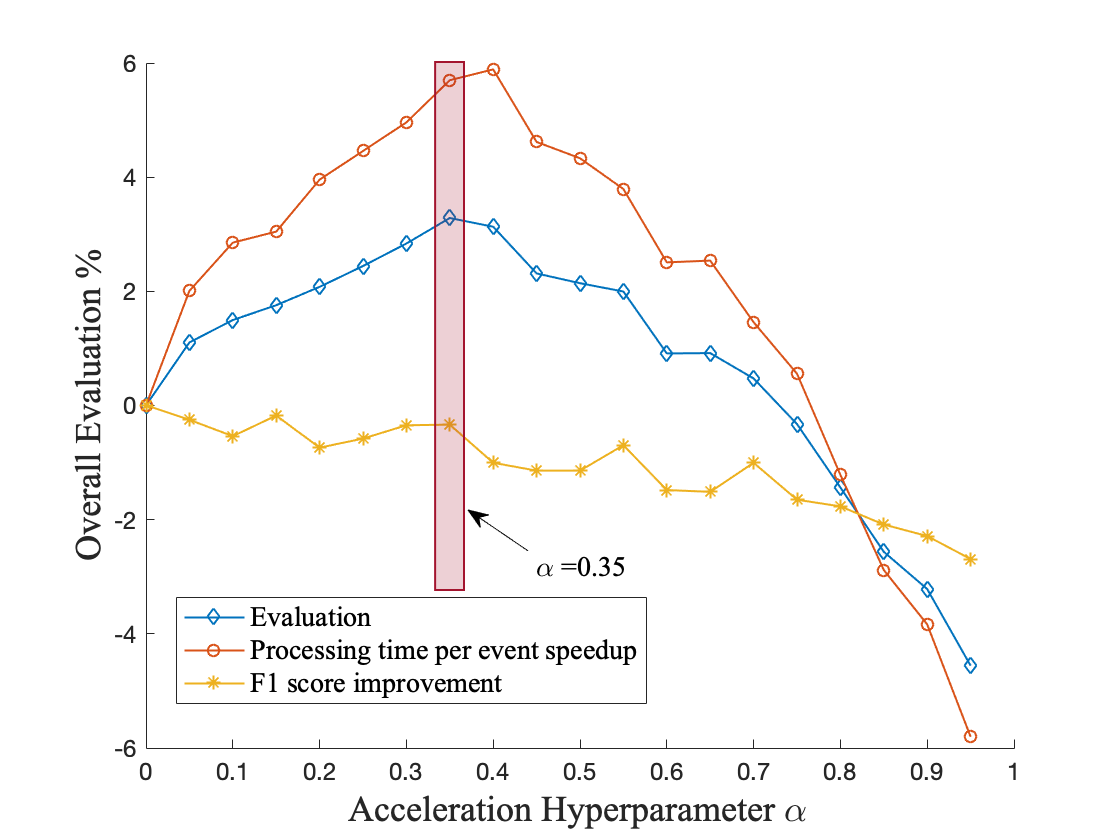}}
    \caption{The relationship between E-score and $\alpha$ value}
    \label{fig:alpha-evaluation}
\end{figure}

The overall evaluation considering both efficiency and accuracy is computed in Equation \ref{eq:e-score}. Fig. \ref{fig:alpha-evaluation} illustrates the relationship between E-score and $\alpha$ value, that our MEMS with strategy-1 performs better than the standard MS when $\alpha <0.75$, especially at $\alpha=0.35$.


\subsubsection{Strategy 2}
\label{subsec:strategy2}

To accelerate MEMS algorithm, one of the most intuitive approaches is to reduce the processing data. Building on that, the hyperparameter $\beta$ is introduced to downsample the original events expressed in Equation \ref{eq:original events} evenly. The downsampled events are represented in Equation \ref{eq:downsampled events}, that the size will be reduced to $1/\beta$ of the original size. 

\begin{equation}
    \textit{Original events: }\textbf{X}=	\left\{ \textbf{x}_1, \textbf{x}_2, ... , \textbf{x}_i 	\right\}
    \label{eq:original events}
\end{equation}
\begin{equation}
    \textit{Downsampled events: }\textbf{X}^\prime=	\left\{ \textbf{x}_1, \textbf{x}_{2 \beta}, ... , \textbf{x}_{n\beta}	\right\}, n\beta \leq i
    \label{eq:downsampled events}
\end{equation}

\begin{figure*}[!h]
    \centering
    \hspace{-1.2cm}
    \begin{subfigure}[t]{0.6\textwidth}
    \centerline{\includegraphics[height=1.9in]{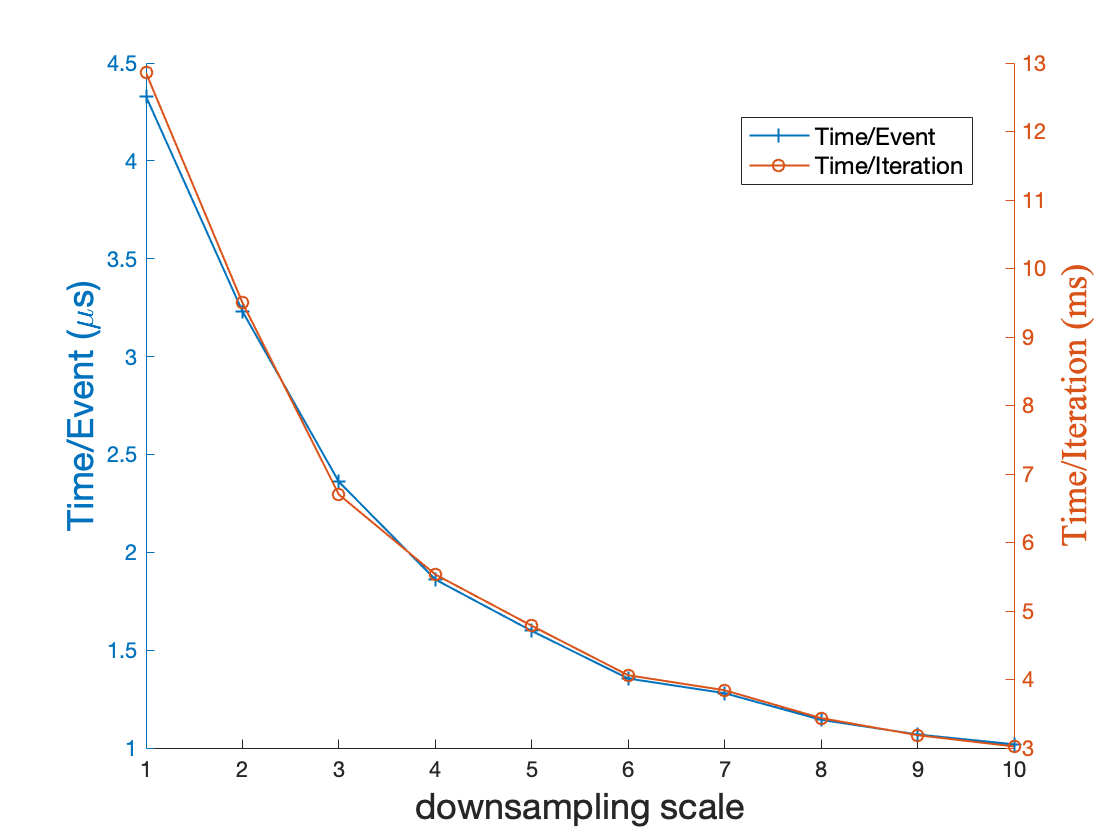}}
    \caption{Processing time}
    \end{subfigure}
    ~
    \begin{subfigure}[t]{0.4\textwidth}
    \centerline{\includegraphics[height=1.9in]{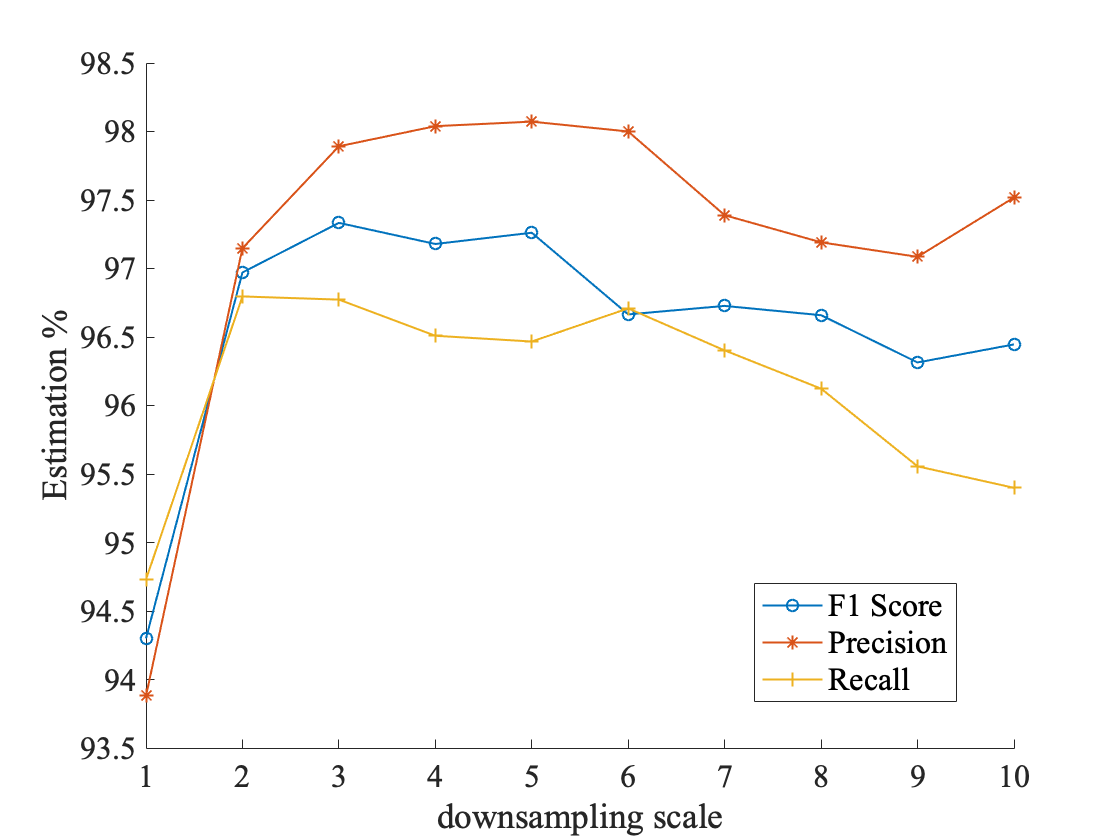}}
    \caption{Performance}
    \end{subfigure}
    
    \caption{Processing time and performance with varying downsampling rate $\beta$}
    \label{fig:beta-event}
\end{figure*}

Similarly, the processing time of each event and mean shift iteration are computed to assess the efficiency, as illustrated in Fig. \ref{fig:beta-event}.

\begin{figure}[!ht]
    \centering
    \centerline{\includegraphics[height=2.2in]{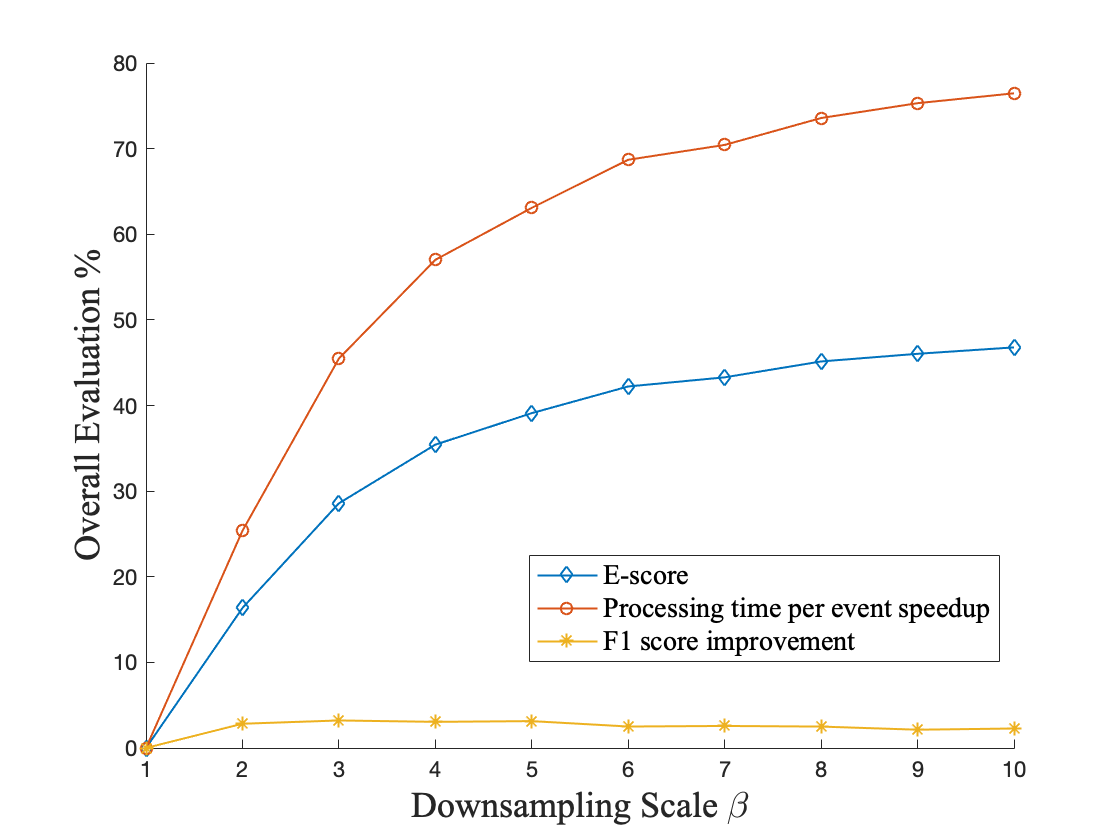}}
    \caption{Overall evaluation with varying downsampling scale $\beta$}
    \label{fig:beta-score}
\end{figure}

From Fig. \ref{fig:beta-event}(a), the processing time evaluated on both single event and iteration demonstrates a decaying trend with increasing $\beta$. The smaller $\beta$ results in the greater reduction slop which means a more significant improvement of efficiency. When $\beta$ exceeds 4, the processing time per event can even reach around $1 \mu s$. 
Fig. \ref{fig:beta-event}(b) presents the assessment of the clustering accuracy, where recall, precision and F1 score outperforms the standard MS as $\beta=1$. With the reduction of original events, the noise captured and included in the original data will also be reduced. As a result of the reduced influence of noise disruption, clustering and segmentation will function more accurate. 
Based on the processing time per event and F1 score, the overall evaluation $E-score$ of MEMS with strategy-2 is calculated with varying $\beta$ values as illustrated in Fig. \ref{fig:beta-score}. It shows a rising improvement with increasing $\beta$ values, and at least around $16 \%$ improvement is reached when $\beta=2$.

\subsection{Visual Servoing}
\label{sec: vs-free}
The traditional Visual Servoing (VS) is based on the information extracted from standard vision sensors, which was first proposed by the SRI International Labs in 1979 [132]. 
Distinguishing to the eye-to-hand configuration relying on observing the absolute position of the target and the hand, an eye-in-hand camera is attached on hand and observes the relative position of the target. The major purpose of VS in this study is to manipulate the gripper to the desired position of the target's centroid in a 2D plane. 
An Event-based Visual Servoing (EVS) method for multiple objects adopting depth information with the eye-in-hand configuration is proposed based on our previous work \cite{muthusamy2021neuromorphic}. The centroid information obtained by the proposed MEMS will guide EVS to track the object. Then the robust corner is further calculated using a heatmap to ensure a stable manipulation. 

\begin{figure*}[!h]
    \centering
    \begin{subfigure}[t]{0.208\textwidth}
        \centering
        \includegraphics[height=1in]{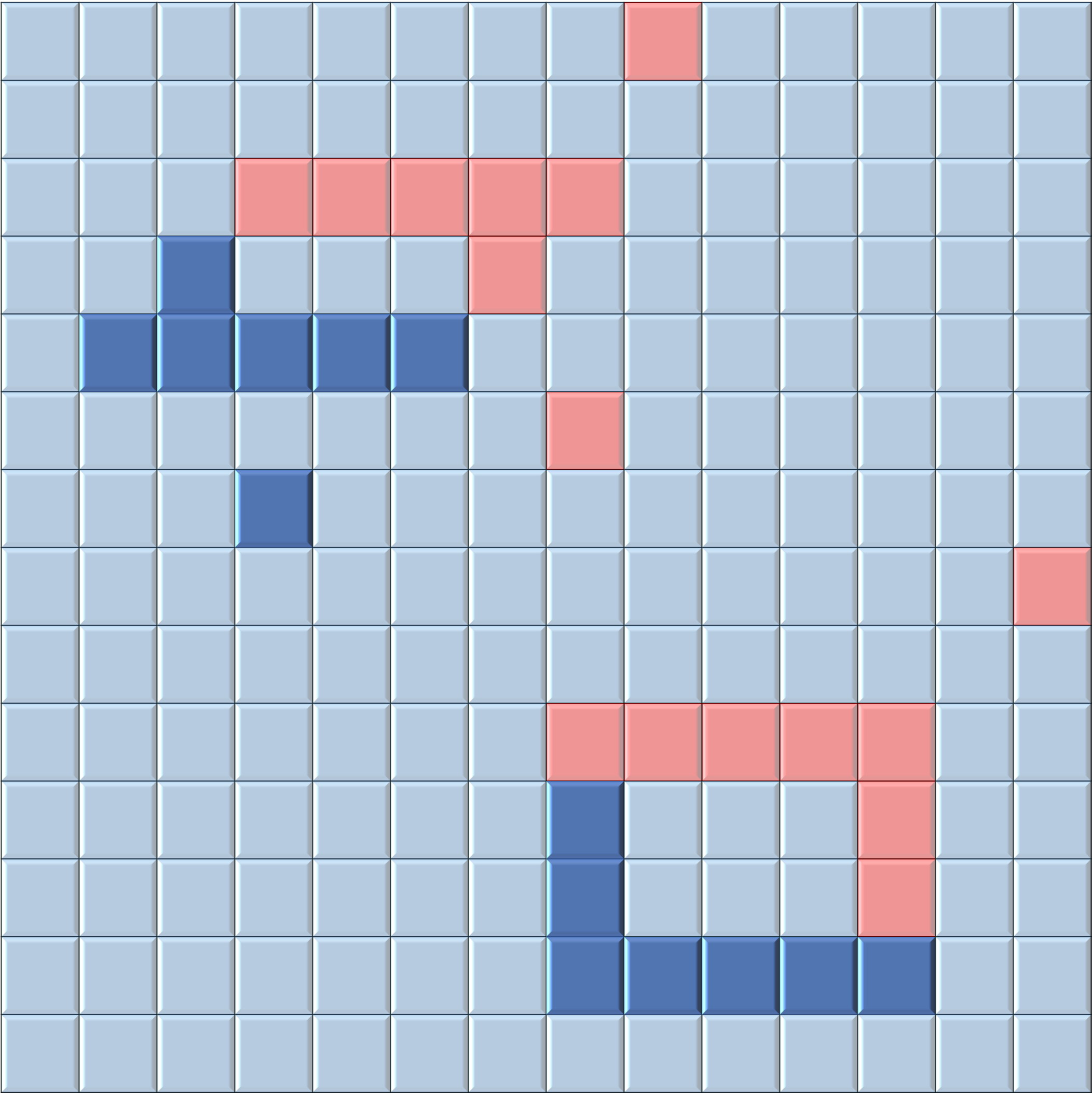}
        \caption{SAE}
    \end{subfigure}%
    ~ 
    \begin{subfigure}[t]{0.2\textwidth}
        \centering
        \includegraphics[height=1in]{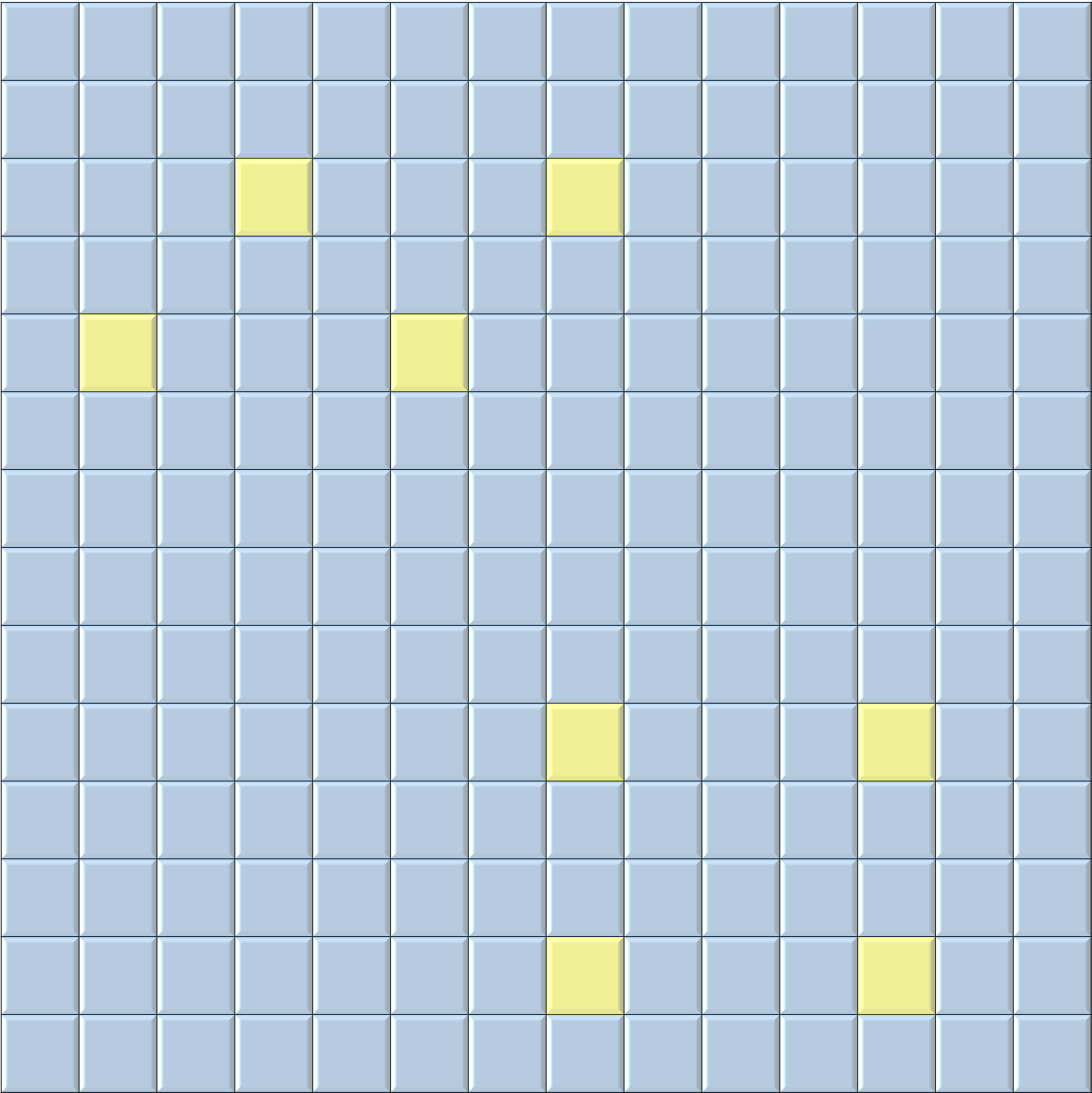}
        \caption{SAFE}
    \end{subfigure}
    ~ 
    \begin{subfigure}[t]{0.2\textwidth}
        \centering
        \includegraphics[height=1in]{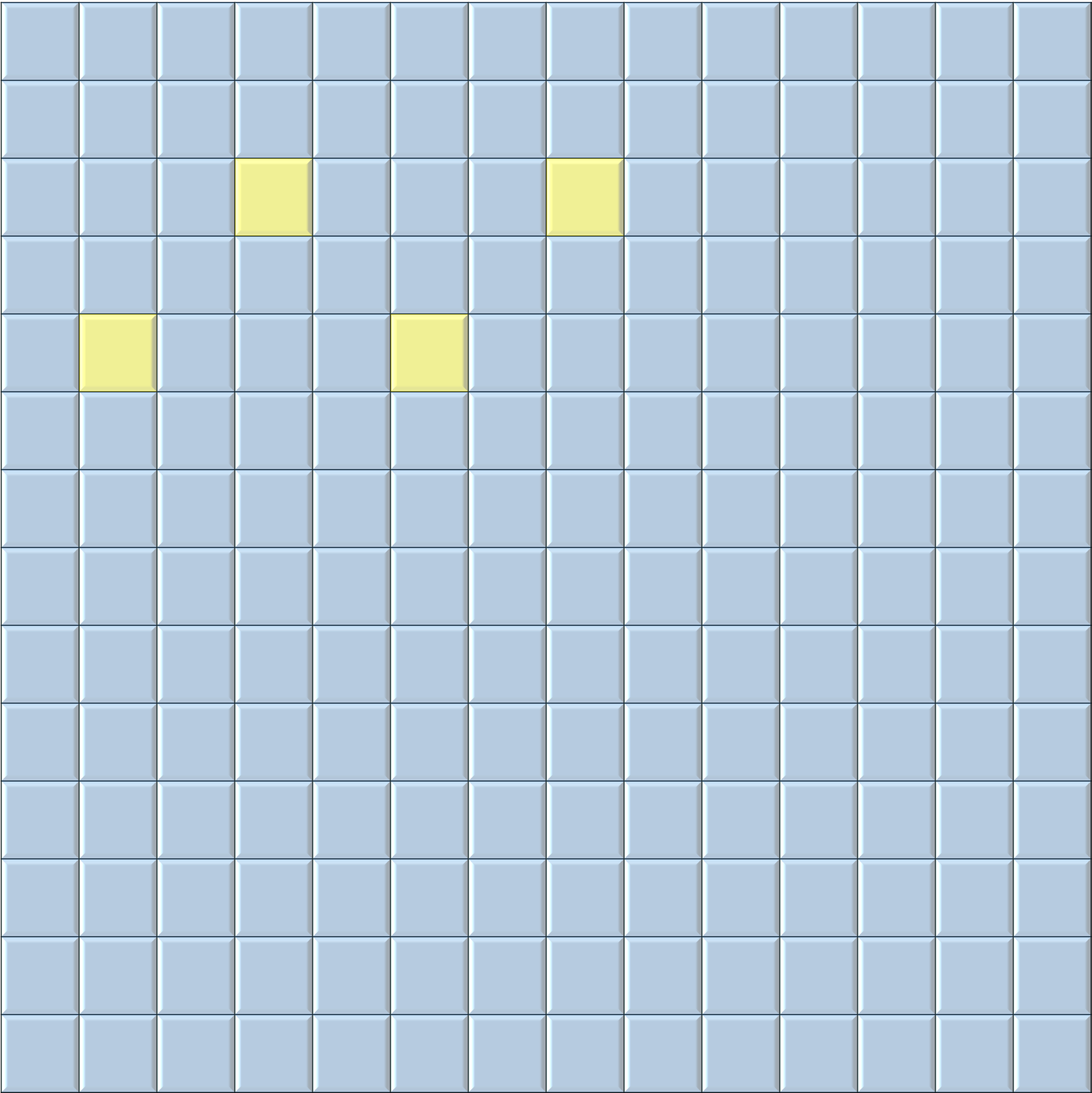}
        \caption{SALE}
    \end{subfigure}
    ~ 
    \begin{subfigure}[t]{0.2\textwidth}
        \centering
        \includegraphics[height=1in]{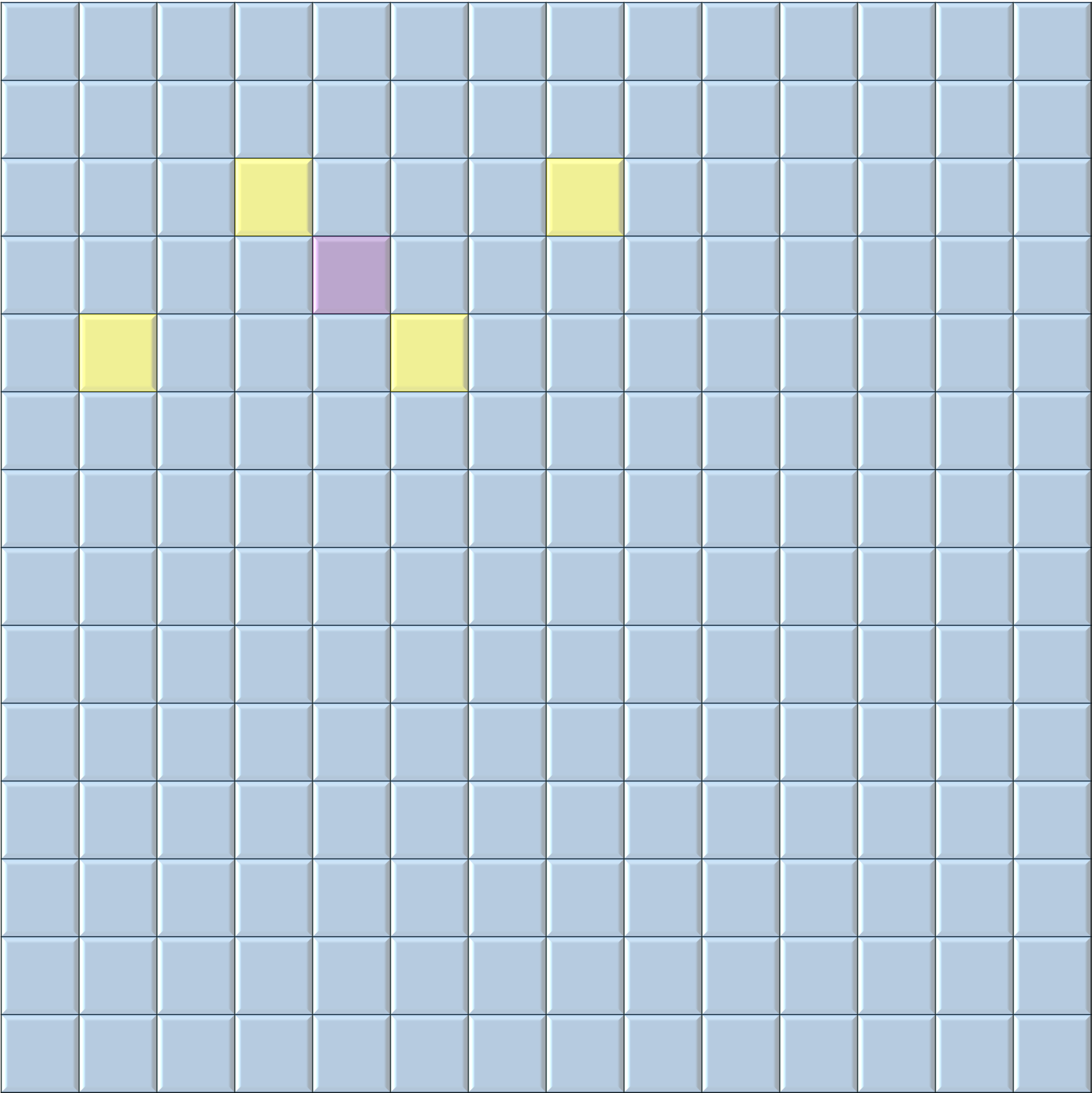}
        \caption{SAVE}
    \end{subfigure}
    \caption{Four layers of surfaces of active events for robust centroid detection. (a) Raw events on SAE. Red and blue blocks represent negative and positive polarity events. (b) Corner features (yellow blocks) on SAFE. (c) Corners of targeting object on SALE. (d) Virtual robust centroid (purple block) on SAVE.}
    \label{fig: SAEs}
\end{figure*}

Four layers of active events surfaces are considered as shown in Fig. \ref{fig: SAEs}. The Surface of Active Events (SAE) shows all the raw events captured by the event-based camera. By using a feature detector, only the corners will be extracted and projected into Surface of Active Feature Events (SAFE). In this work, eHarris detector is applied to detect corner features of the objects. 
The mask is applied to remove the corners of other objects, so only the useful features are remained and projected to the Surface of Active Locking Events (SALE). The robust centroid information will be calculated and virtually projected into the Surface of Active Virtual Events (SAVE). 

\begin{figure}[!h]
    \centering
    \centerline{\includegraphics[scale=0.15]{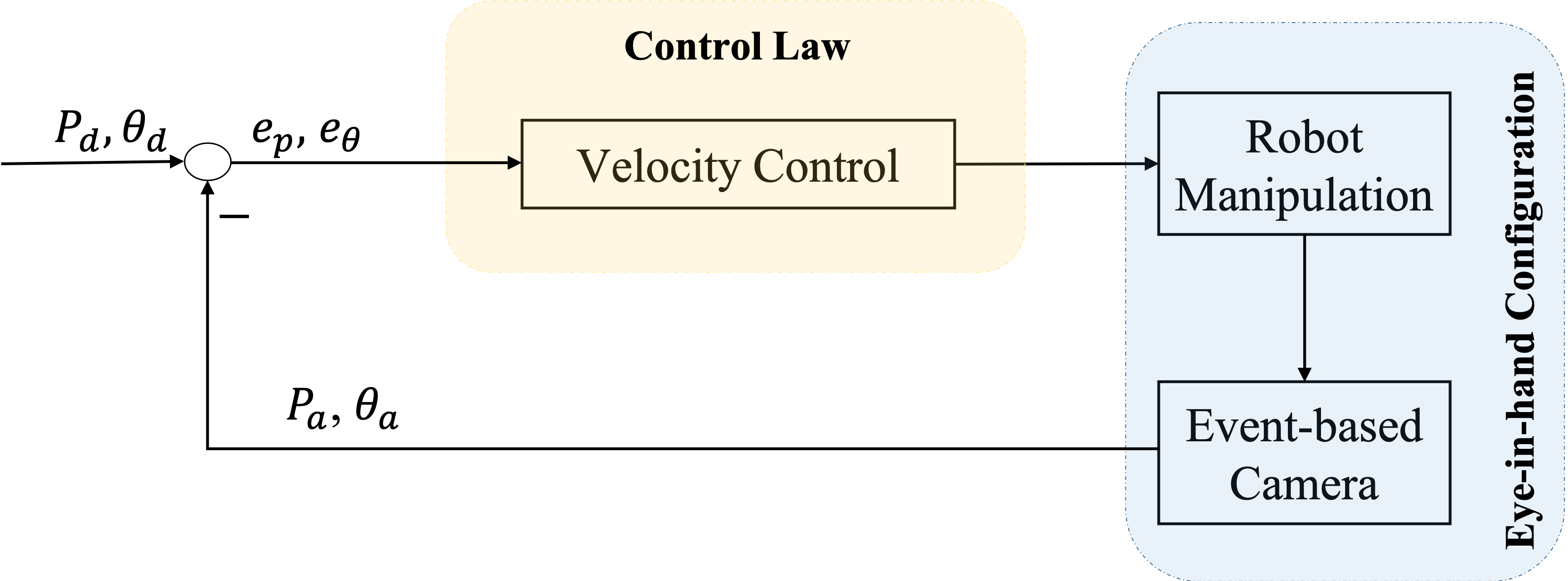}}
    \caption{The block diagram of event-based visual servoing.}
    \label{fig:vs}
\end{figure}

According to the robust centroid and depth information, the robot will be manipulated to track the object. The block diagram for completing the manipulation task by EVS is depicted in Fig. \ref{fig:vs}. 
$P_d$ and $P_a$ represent the desired and the actual/current planar position of the object’s centroid. $\theta_d$ and $\theta_a$ indicates the desired and the actual/current orientation of the object. The error $e_p$ and $e_\theta$ can be calculated as $P_d-P_a$ and $\theta_d-\theta_a$. 
According to the position error, a forward and lateral correction will be executed to move the gripper to the proper position. Then the angular correction will be implemented in order according to the angular error. Based on the control and manipulation, the orientation and position error will be eliminated until the robot aligns with the object.  

\begin{figure*}[h!]
    \centering
    \centerline{\includegraphics[scale=0.25]{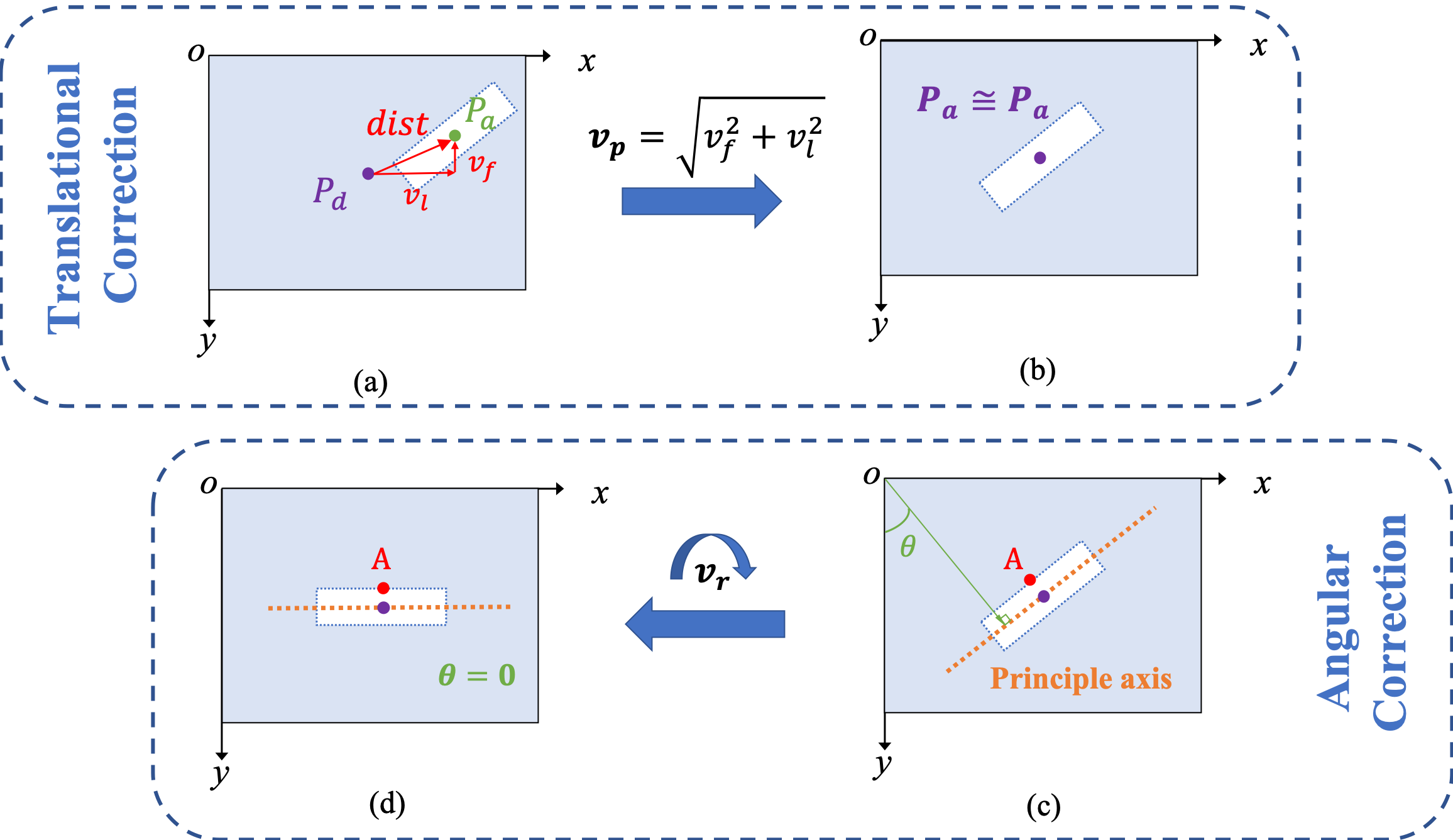}}
    \caption{Velocity control principle of EVS which includes two parts in order: translational correction (a-b) and angular correction (c-d). }
    \label{fig:vs_gg}
\end{figure*}

The UR robot utilized in this work provides the secondary velocity control $\Vec{v}(\Vec{v}_f, \Vec{v}_l, \Vec{v}_r)$ for end-effectors/grippers, and the relationship between the moving distance $dist$ and planar velocity $\Vec{v}_p$ is as $dist=|f(\Vec{v}_p)|$. Here $\Vec{v}_f, \Vec{v}_l$ and $\Vec{v}_r$ represent the forward, lateral and rotational velocity, respectively. Besides, the planar velocity can be computed as $\Vec{v}_p=\Vec{v}_f+\Vec{v}_l$. The velocity control based visual servoing is illustrated in Fig. \ref{fig:vs_gg}, that consists of two stages - translational (forward and lateral) correction and angular correction in order. 

Fig. \ref{fig:vs_gg} (a-b) show the sequence of translational correction, and the estimated position error $dist$ in image coordinate will be eliminated by velocity $\Vec{v}_p(\Vec{v}_f+\Vec{v}_l)$ until $dist$ is smaller than the threshold. 
For Barret hand with three fingers, the grasp hypothesis generated is required to ensure a proper and stable grasp. In this work, grasp is planned based on the principal orientation which will be elaborated in Section \ref{subsec:grasp plan}. 
As depicted in Fig. \ref{fig:vs_gg} (c-d), the gripper will keep rotating with $\Vec{v}_r$ until the angle difference $\theta$ is eliminated to near $0$. After accomplishing correction, the gripper will move down to pick the object up according to the depth information mapped. 

\subsection{Grasping Plan}
\label{subsec:grasp plan}

For model-free object grasping, there is no geometric model or any prior knowledge of the object. A mount of researches rely on exploring the geometrical information such as shapes, edges, and saliency. It suffers from low efficiency, since the exploration by moving the camera around the unknown objects takes time. Another popular approach is using deep learning techniques such as DCNN to train robots to generate a proper grasping hypothesis but it requires a vast amount of manually labeled data for training. 
Therefore, a fast grasp generation is proposed for unseen objects using Principal Component Analysis (PCA), which is relatively more efficient by avoiding online exploration and offline training.

In this work, the principal axis of objects obtained from PCA is utilized to generate a proper grasp position. The principal component is equivalently defined as a direction that maximizes the variance of the projected data, which can be computed by eigen decomposition of the covariance matrix $COV$ of the data matrix as described in the following equation:
\begin{equation}
    COV=
    \begin{pmatrix}
        \sigma^2_{xx} & \sigma^2_{xy} \\
        \sigma^2_{yx} & \sigma^2_{yy} 
    \end{pmatrix}
\end{equation}
where $\sigma^2_{xx}, \sigma^2_{xy}, \sigma^2_{yx}$ and $\sigma^2_{yy}$ are the covariance values of 2D coordinate. 
It is based on calculating the eigenvalues ($\lambda_1 > \lambda_2$) and the corresponding eigenvectors ($u_1, u_2$) to find the principal component, where eigenvectors and eigenvalues are used to quantify the direction and the magnitude of the variation captured by each axis. 
Then $u_1 (u_{1x}, u_{1y})$ can approximate the direction $\theta$ of the principal axis as: 
\begin{equation}
    \theta=arctan\frac{u_{1y}}{u_{1x}}
\end{equation}
The grasping pose will be generated by the centroid $C$ and the direction $\theta$. 
To ensure a robust principal orientation, all the orientations detected before grasping will be stored in a histogram with 3-degree bins. The final rotation is converted into the range of $[-90, 90]$ to ensure the shortest rotation path, resulting in 61 bins in the histogram. 
Then the final robust orientation is determined by the bin value with the maximum probability in the histogram.

\begin{figure*}[!ht]
    \centering
    \includegraphics[height=1.2in]{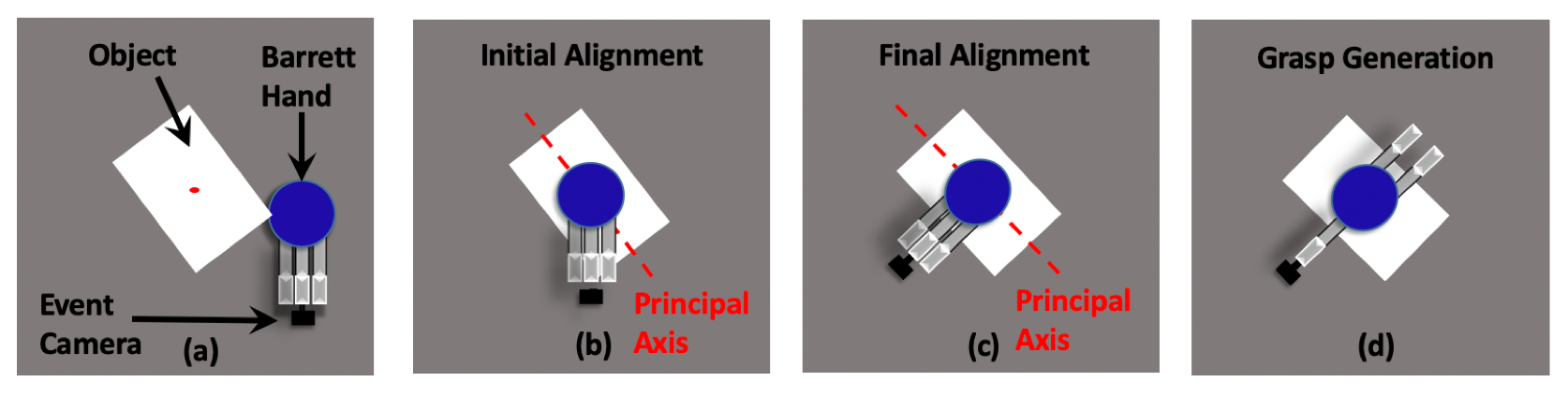}
    \caption{Barrett hand grasp alignment }
    \label{fig: BHGraspAlignment}
\end{figure*}

The Barrett hand used in this work has three fingers of eight joints with only four degrees of freedom. Each finger contains two degrees of freedom controlled by a servo-actuated joint. Two of the fingers have an extra joint which allows them to rotate synchronously around the palm with a certain spread angle relative to the third finger up to 180 degrees. The three fingers are commanded with the same joint value, simplifying the grasp plan and limiting the number of possible configurations. Fig. \ref{fig: BHGraspAlignment} shows the knowledge-based approach that is used in our case to find the appropriate grasp plan, 1) the grasping point on the object (centroid), 2) the principal axis of the object. To perform a grasp, a Tool Center Point (TCP) is defined first on the Barrett hand. The hand is moved to the grasping point on the object and rotate to be perpendicular to the object's principal orientation. 
Next, the fingers are closed around the object until contacts or joint limits prevent further motion. This configuration is executed after ensuring that the Barret hand moving fingers can achieve stable contact points with the object's side surfaces. The distance between the two moving fingers is pre-measured and compared with the edge length of the side surface to confirm the grasp. 

\subsection{Event-based Multiple Objects Grasping Framework}

\begin{figure}[h!]
    \centering
    \centerline{\includegraphics[scale=0.12]{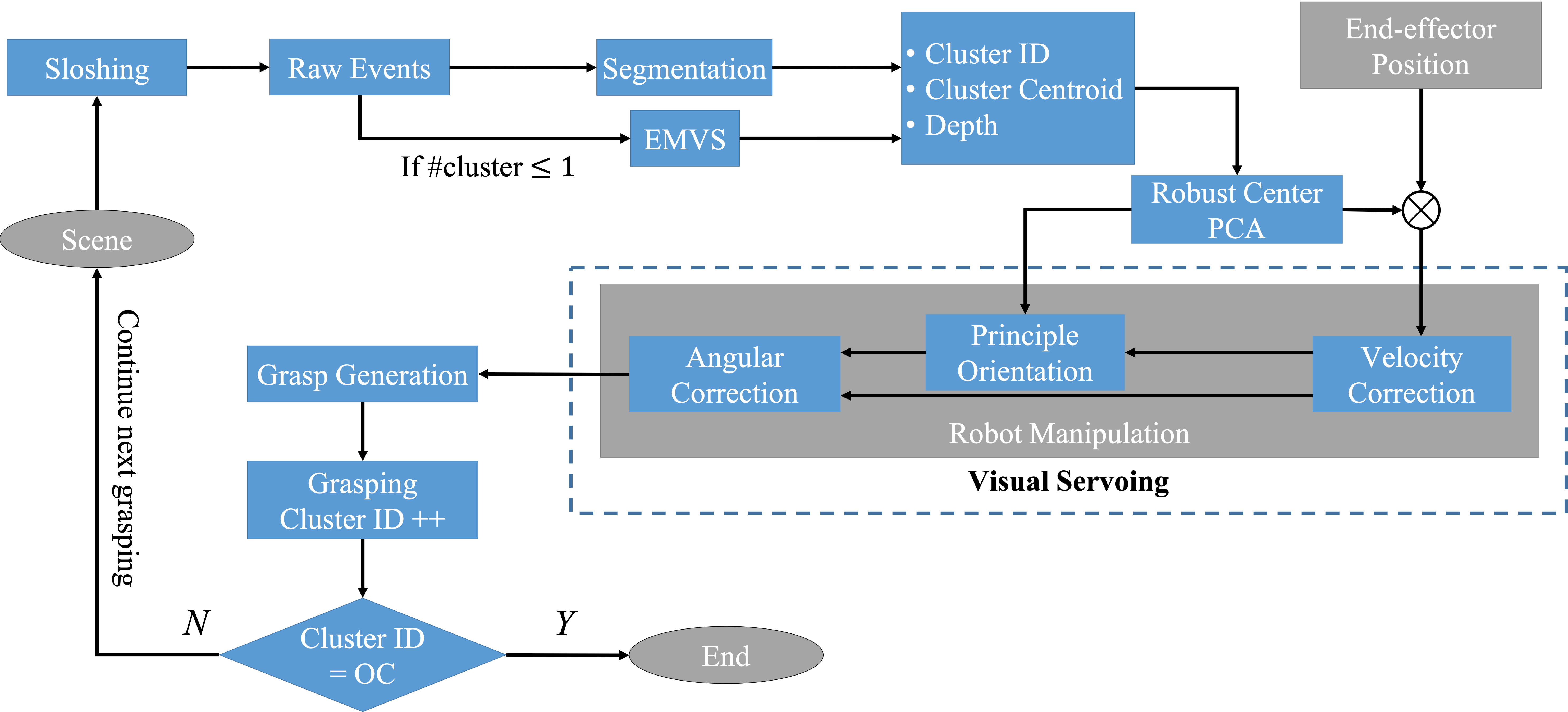}}
    \caption{The whole framework of proposed purely event-based multiple-object grasping.}
    \label{fig:mul-pipeline}
\end{figure}

The whole framework of the proposed model-free neuromorphic vision-based multiple-object grasping is illustrated in Fig. \ref{fig:mul-pipeline}. 

In this approach, each object will be numbered from one after segmentation and grasped orderly according to their IDs. The depth information obtained by EMVS will be used only once before grasping, since the additional movement and time are required and consumed. Thus the one-time EMVS will be executed in the initial stage before segmentation when the cluster ID is equal to one, and the depth map of the initial position will be frozen. By using the developed MEMS, objects will be clustered with their IDs and centroid position.  Through mapping the frozen depth map with the segmented object information, the 3D spatial information of each object at the initial position will be obtained and locked to provide the depth information for the next grasping. As depicted in Section \ref{sec: vs-free}, the robust centroid of the current tracking object will be obtained and projected into SAVE. Based on the position error and orientation error, the translational and angular correction will be employed until the gripper is aligned to the targeting object. Then the object will be picked up and placed at the specific dropping area. After that, the cluster ID will be accumulated by 1, and the gripper will return to the initial position and start the next grasping task. 
The framework of multiple-object grasping is summarized in Algorithm \ref{algo:multi-obj grasp}.
\begin{figure}[!h]
\begin{algorithm}[H]
\DontPrintSemicolon

  \KwInput{Events stream: position ($x_i, y_i$), polarity $p_i$, timestamp $t_i$}
  \KwOutput{Cluster ID, cluster centroid ($x_c, y_c$)}
Initialize cluster ID = 1 \\
Set starting point of gripper $P_0$\\
\While{ID = 1 or ID $\leq$ The number of objects}{
  \eIf{ID = 1}{
   Move gripper and record the trajectory\;
   Perform EMVS and MEMS\;
   Map and freeze depth with 2D spatial and centroid position of each object\;
   }{
   Perform MEMS to detect the centroid of the targeting object\;
   Obtain depth, 2D spatial and centroid information of each object\;
  }
  Detect corners in SAE by applying e-Harris, and project corner events to SACE\;
  Extract object corners in SACE using heatmaps\;
  Calculate the robust centroid of the targeting object\; 
  Calculate the position error $e_p$ between the current position and robust centroid position\;
  \eIf{$abs(e_p) > 0$}{
  Perform EVS to eliminate $e_p$
  }{
  Perform EVS to eliminate the angular error $e_r$\;
  Execute grasping\;
  }
  \If{grasp accomplished}{
   Cluster ID +1\;
   Move back to the initial position\;
  }
  }
\caption{Purely Event-based Multiple-object Grasping }
\label{algo:multi-obj grasp}
\end{algorithm}
\end{figure}

\section{Experimental Validation on Multiple-object Grasping}
\label{sec: experiment}
This section describes the experimental validation of multiple-object grasping and discusses the experimental results.

\subsection{Experimental Setup and Protocol}
\begin{figure*}[!ht]
    \begin{subfigure}[t]{0.4\textwidth}
        \centering
        \includegraphics[height=2.5in]{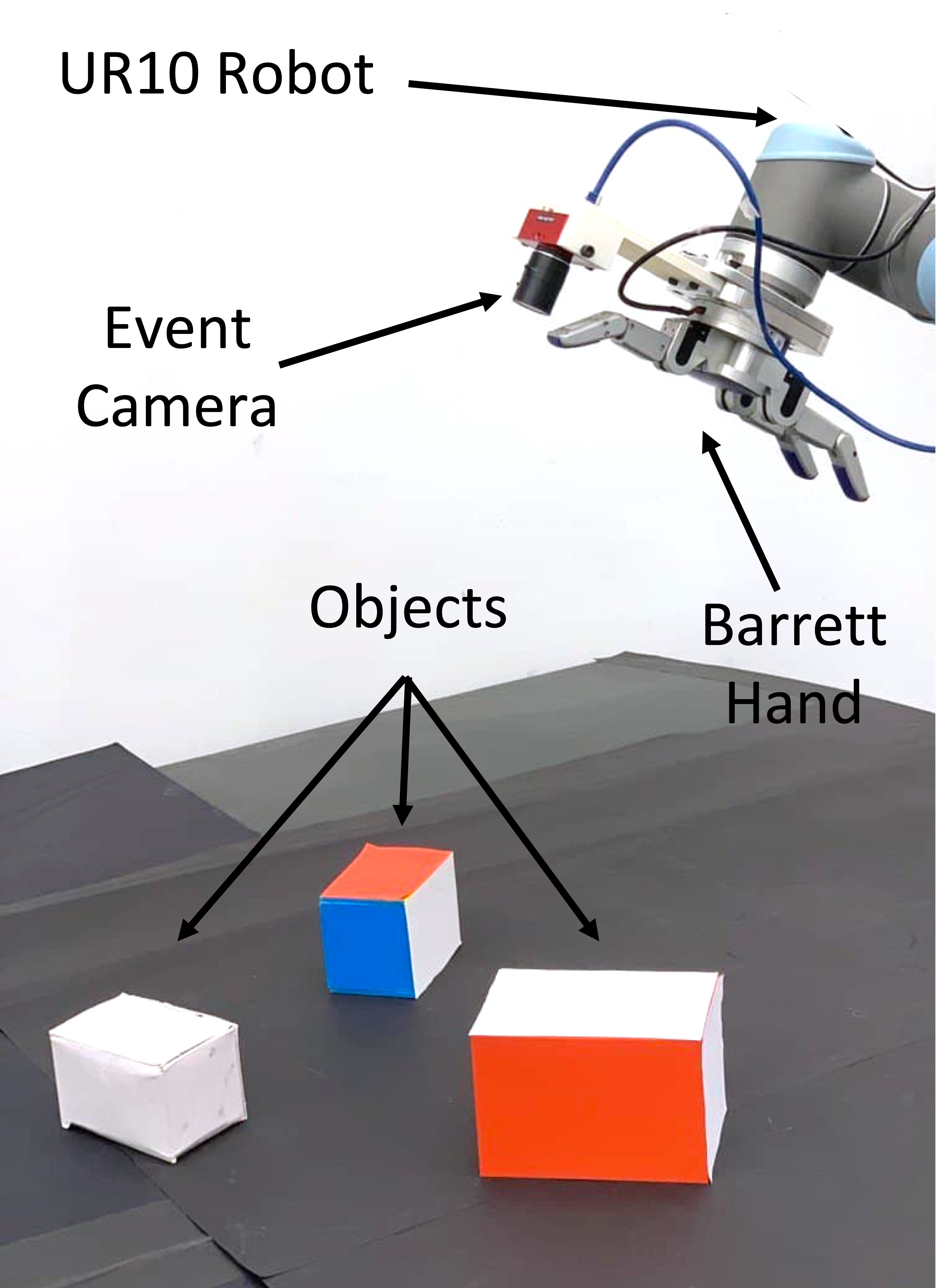}
        \caption{Robotic grasping system}
        \label{fig: setup1}
    \end{subfigure}%
    ~ 
    \hspace*{-0.35cm}  
    \begin{subfigure}[t]{0.4\textwidth}
        \centering
        \includegraphics[height=2.5in]{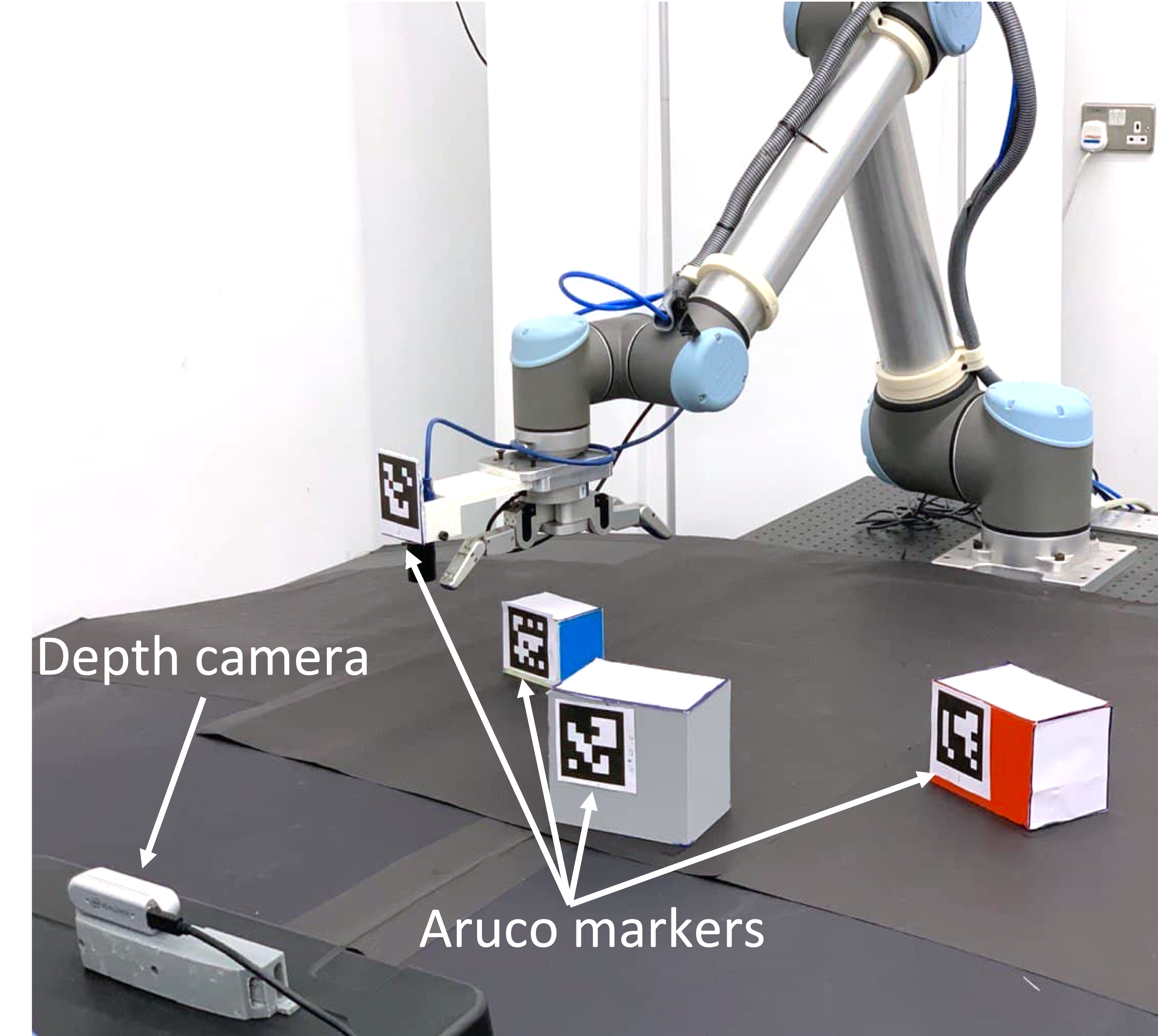}
        \caption{Evaluation system}
        \label{fig: setup2}
    \end{subfigure}
    \caption{Experiment setup consists of two parts: (a) Robotic grasping system for experimental validation of proposed approach; (b) Evaluation system for assessing the grasping performance. }
    \label{fig: setup}
\end{figure*}

The real experiments are performed to validate the proposed grasping approaches. As demonstrated in Fig. \ref{fig: setup}, the experimental setup consists of a robotic grasping system and an evaluation system. 

The grasping system includes a Universal Robots UR10 6-DOF arm \cite{UR10}, a Barrett hand gripper \cite{Barrett}, and a Dynamic and Active pixel VIsion Sensor (DAVIS346) \cite{DAVIS346} placed in an eye-in-hand configuration. The UR10 arm features a $10kg$ weight capacity and $0.1mm$ movement accuracy, making it ideal for packaging, assembly, and pick-and-place tasks. 
The DAVIS346 sensor has an outstanding dynamic range ($>100 dB$) and $346*260$ resolution. The stream of events encodes time $t$, position $(x, y)$ and sign of brightness change $p$. Objects of different sizes and shapes are used as the grasping targets. 
To perform the proposed approach successfully, it is assumed that the targeting objects are within the gripper’s manipulation range since the robot is installed in a fixed base. Moreover, the sizes of objects are expected to be within the maximum opening of the gripper. 

\begin{figure*}[h!]
    \centering
    \begin{subfigure}[t]{0.31\textwidth}
        \centering
        \includegraphics[height=1.8in]{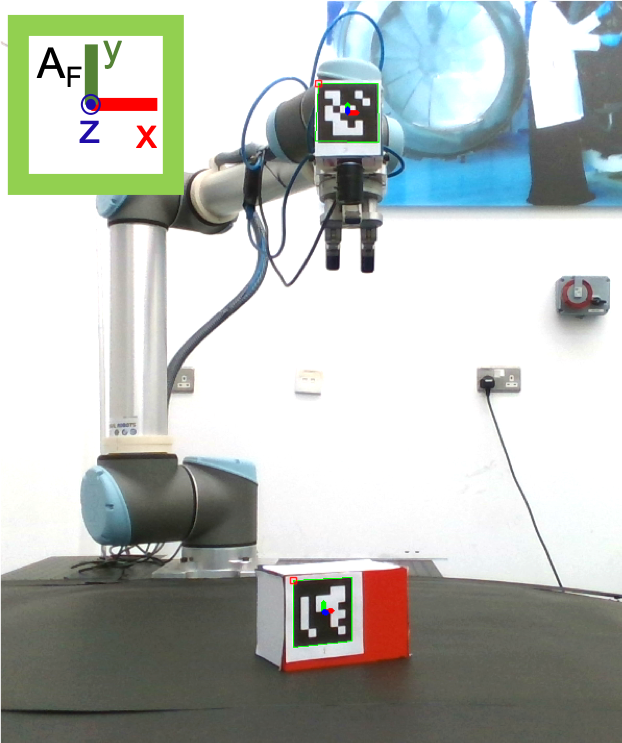}
        \caption{Initial pose of object $P_{obj}$}
        \label{fig: Aruco-a}
    \end{subfigure}%
    ~ 
    \begin{subfigure}[t]{0.31\textwidth}
        \centering
        \includegraphics[height=1.8in]{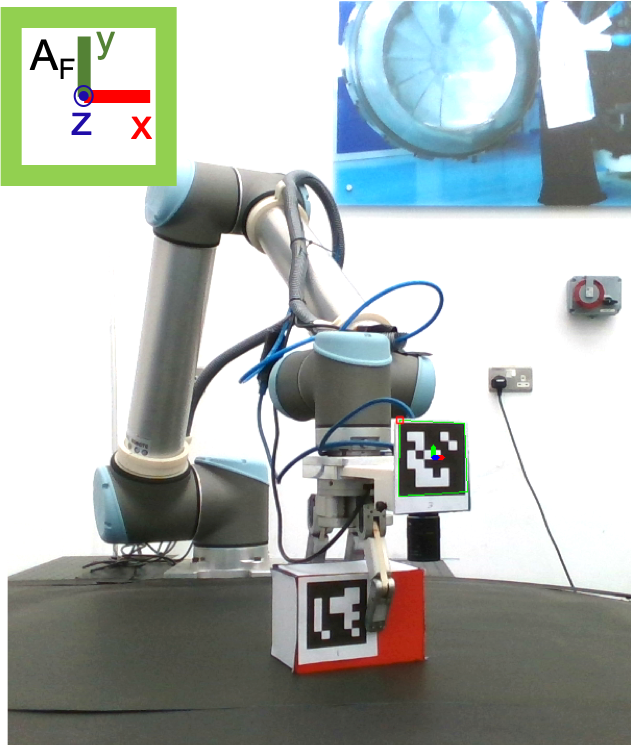}
        \caption{Grasping pose of Barrett hand $P_{grip}$}
    \end{subfigure}
    ~ 
    \begin{subfigure}[t]{0.31\textwidth}
        \centering
        \includegraphics[height=1.8in]{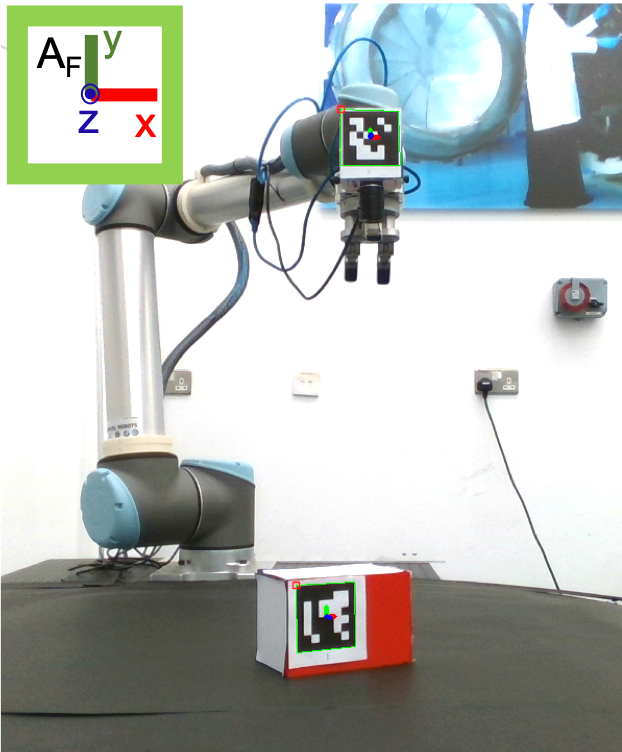}
        \caption{Deviated pose of object $P_{obj}\prime$}
    \end{subfigure}
    \caption{Pose estimation by developed evaluation system in three steps: (a) Initial pose   (b) Optimal grasping pose (c) Deviated object pose after grasping. The coordinate of ArUco markers is indicated at the left top corner. }
    \label{fig: Aruco}
\end{figure*}

To estimate the grasping performance, we developed an evaluation system that consists of ArUco markers 
and a standard camera Intel D435. The identity of an ArUco maker is determined by its binary matrix inside of the black border, that facilitates a fast detection and applicability for camera calibration and pose estimation. By conducting 10 experiments of measuring the static object's pose using our evaluation system, the estimation error of angle and position are evaluated as only $1^\circ$ and $0.1cm$, respectively. In this work, the identified ArUco markers are attached on the lateral side of gripper and targeting objects, so their poses can be determined by detecting and estimating the pose of ArUco markers. According to the evaluation metrics for grasping performance detailed in Section \ref{sec: metrics}, we focus on the poses in three stages: initialization, optimal grasping and object grasping as demonstrated in Fig. \ref{fig: Aruco}. In the beginning, the object's pose will be recorded as $P_{obj}$. After visual servoing, the gripper will reach the optimal grasping pose as $P_{grip}$. Since the ArUco marker would be covered by the finger of Barrett hand after grasping, Barrett will hold the object for one second and release it. After opening the gripper, the object pose will be estimated as $P_{obj}\prime$ to evaluate the object deviation. 

\subsubsection{Model-free grasping experiment protocol}
According to Algorithm \ref{algo:multi-obj grasp}, the experiments are designed and performed in the following steps:\\
1) Depth exploration stage. Move the gripper in a linear trajectory to the initial position and perform EMVS to obtain the depth information. This stage is only activated once at the start of the whole experiment. \\
2) Segmentation stage. Slosh gripper to generate some movement for observing the objects, since only illumination change can be captured by the event camera. Then segment each object by the developed MEMS to obtain the centroid information. 
Meanwhile, the orientation of each object is acquired by PCA. Sort objects according to volume, and update centroid and orientation information of the largest object to visual servoing. \\ 
3) Visual servoing stage. Extract the robust corner feature and virtual object centroid in SAVE, and track the object until the object and camera centers are matched. \\
4) Optimal grasping stage. Rotate the Barrett hand to align to the object, and adjust the gripper's position to compensate the installation deviation between the camera and gripper. After rotation and adjustment, Barrett hand will reach the grasping point and hold the object. \\
5) Pick and place stage. Barrett hand lifts and places the object into the drop box in this phase.


\subsubsection{Model-based grasping experiment protocol}
The model-based grasping framework shown in Algorithm \ref{alg:model-based}, shows that the grasping approach is divided in the following steps:\\
1) Scene scanning stage: the robot end-effector starts from a known point and scans the scene in a linear trajectory set of movements.\\
2) Object Localization stage: detected objects' corners are used as an input for the event-based multi-view localization approach, and the objects are localized. \\
3) Point cloud processing stage: point cloud downsampling and object Euclidean clustering is performed to divide the objects in the scene to separate point clouds.\\
4) Model registration stage: for each object an inexact model is fitted to to the detected objects, and the transformation matrix is extracted.\\
5) Grasping stage: The robot gripper is navigated towards the object using PBVS and grasp is performed with the required manipulation.

\subsection{Evaluation Metrics}
\label{sec: metrics}
Proper metrics are crucial to quantify the grasping quality and evaluate the performance of real grasping. In this work, the accuracy of grasping is assessed by the position and orientation error in two phase: optimal grasping and object deviation evaluations. Building on that, the success rate of grasping pose and the grasping quality score are computed to indicate the overall grasping performance. 
\subsubsection{Optimal Grasping Evaluation}
\begin{figure}[!h]
    \centering
    \centerline{\includegraphics[scale=0.2
]{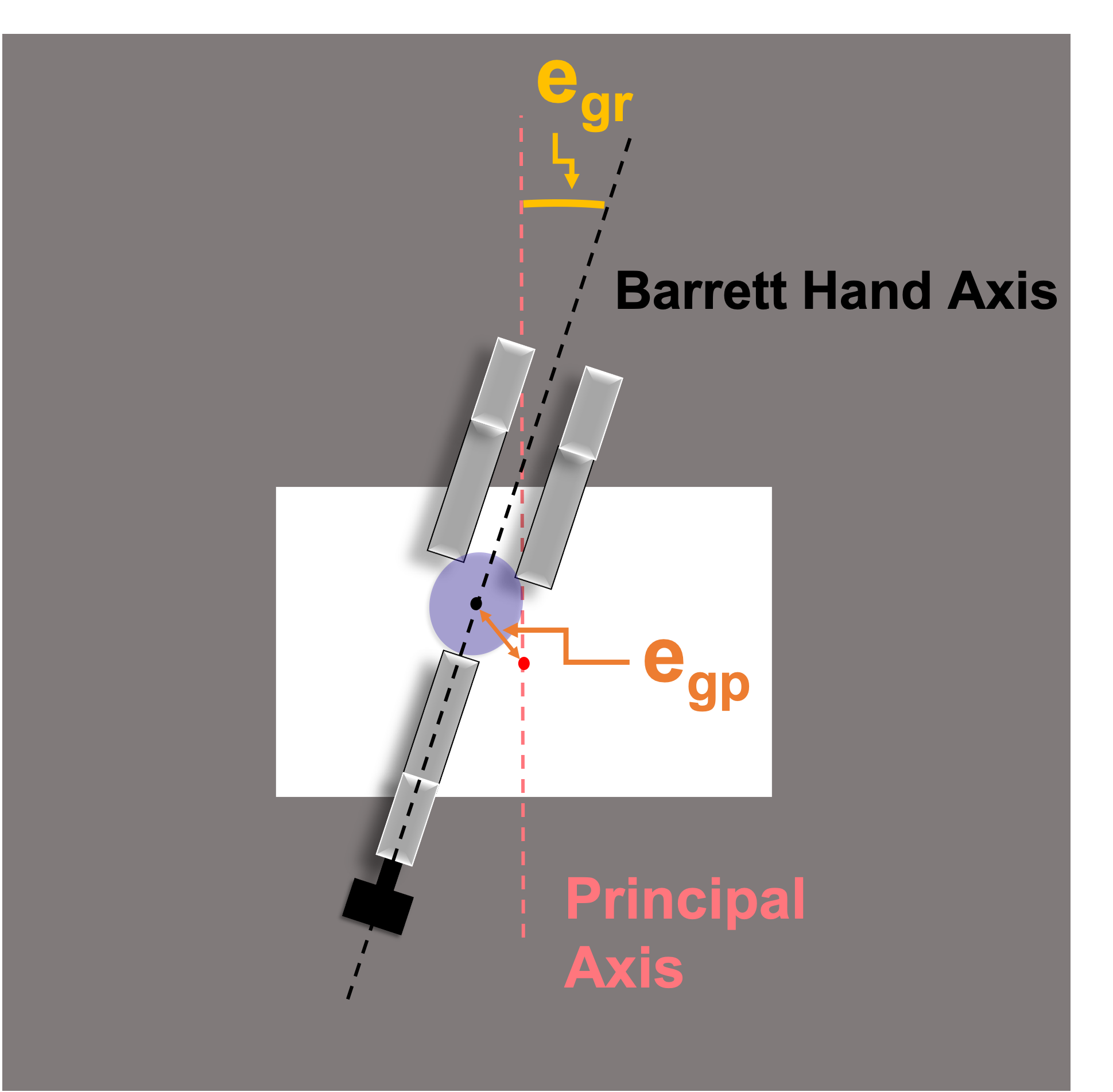}}
    \caption{Two metrics of grasping performance evaluation--positioning error $e_{gp}$ and angular error $e_{gr}$.}
    \label{fig: metrics}
\end{figure}

The goal of optimal grasping evaluation is to meaure the difference between the optimal grasping pose and the actual grasping pose of the gripper after the alignment and before enclasping the object, using two components as illustrated in Fig. \ref{fig: metrics} including the position error $e_{gp}$ and the orientation error $e_{gr}$ of the planning grasp pose. 

The position error $e_{gp}$ and the orientation error $e_{gr}$ represent the distance and the angle between the gripper's center and the actual object's center, respectively. 
We set the limitation of position error $L_{P}$ and orientation error $L_{R}$ to $2cm$ and $15 ^{\circ}$.
Only when both grasping errors are within the limitations, the grasping can be considered as successful as described in Equation \ref{eq:ss}, where $SS$ indicates the success sign of the current grasping. Then the overall success rate can be computed as $R=\sum_N SS/N$, 
where $N$ denotes the total number of grasping performed. 
\begin{equation}
  SS = %
  \begin{cases}
    0 & \text{if ($e_{gp}$ $\leq$ $L_{P}$ and $e_{gr}$ $\leq$ $L_{R}$)}\\
    1 & \text{if ($e_{gp}$ $>$ $L_{R}$ or $e_{gr}$ $>$ $L_{R}$)}
  \end{cases}
  \label{eq:ss}
\end{equation}


\subsubsection{Object Deviation Evaluation}
However, the overall grasping quality can not be estimated only using the planed grasping error before the real grasping. Then the deviation of object pose $D$ is taken into account, reflecting the relative pose before ($\Vec{P}_{b}, R_{b}$) and after ($\Vec{P}_{a}, R_{a}$) trapped by the fingertips of the gripper. The deviation can be quantified as two parts: the position deviation $D_P$ and the orientation deviation $D_R$ as expressed in Equation \ref{eq:D}.
\begin{align}
\begin{split}
 D = \{D_P, D_R\}
\\
 D_P= \|\Vec{P}_{b} - \Vec{P}_{a}\|
\\
 D_R= |R_{b} - R_{a}|
\end{split}
\label{eq:D}
\end{align}

The grasp quality score $Q_{G}$ is calculated according to the deviations and the predefined limitations as expressed in Equation \ref{eq: grasp quality}. Grasping with less object deviation is considered to be of better quality, as the deviation of the object's pose would cause grasping failure.   
\begin{equation}
    Q_{G} = %
  \begin{cases}
    1-\frac{D_P}{2*L_{P}}-\frac{D_R}{2*L_{R}} & \text{if ($D_P$ $\leq$ $L_{P}$ and $D_R$ $\leq$ $L_{R}$)}
    \\
    0 & \text{if ($D_P$ $>$ $L_{P}$ or $D_R$ $>$ $L_{R}$)}
  \end{cases}
  \label{eq: grasp quality}
\end{equation}

\subsection{Experimental Results and Analysis}
\label{sec: results}
\begin{figure*} []
\centering
\rotatebox{90}{Depth Exploration Stage}
\subfloat[EMVS]{\includegraphics[width=0.4\textwidth, height=0.3\textwidth]{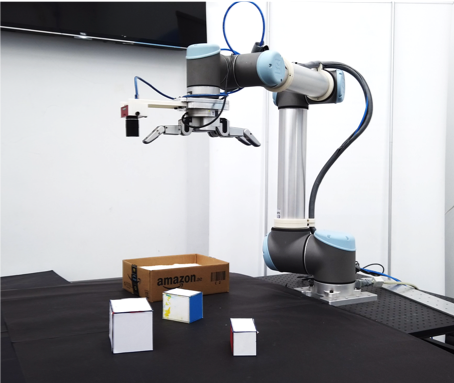}}\label{a}\hfil
\subfloat[Top view of the table]{\includegraphics[width=0.4\textwidth, height=0.3\textwidth]{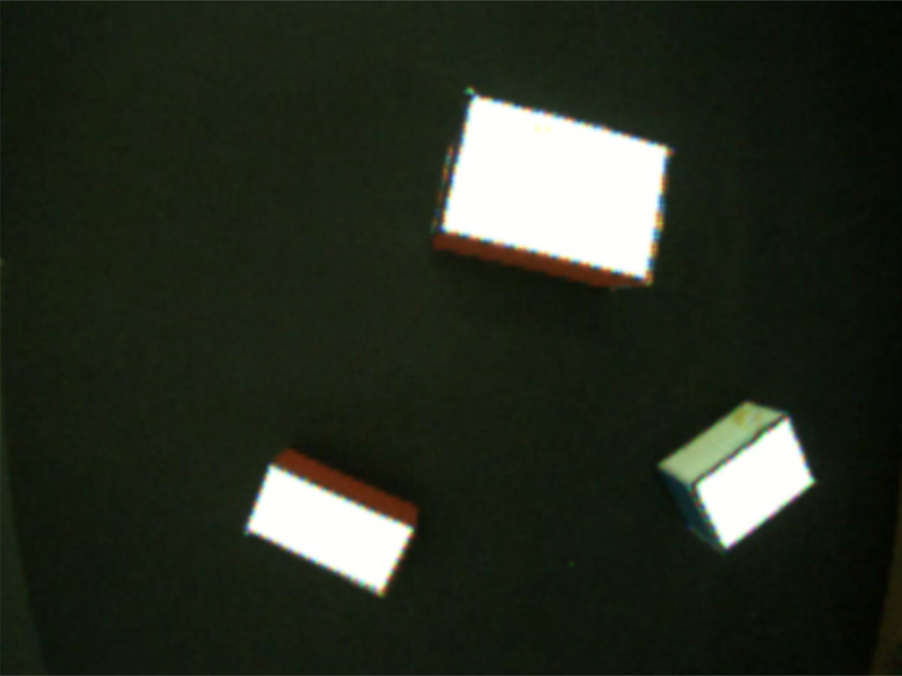}}\label{b}\\
\rotatebox{90}{Segmentation Stage}
\subfloat[Sloshing and segmentation]{\includegraphics[width=0.4\textwidth, height=0.3\textwidth]{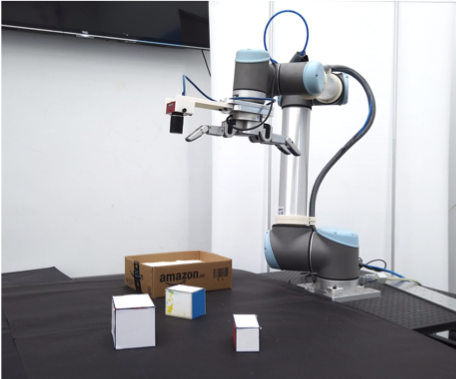}}\label{c}\hfil
\subfloat[Robust corners in SAFE]{\includegraphics[width=0.4\textwidth, height=0.3\textwidth]{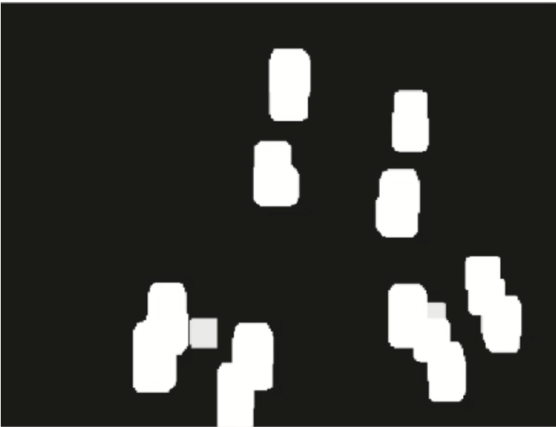}}\label{d}\\
\rotatebox{90}{Visual Sevoing Stage}
\subfloat[Visual servoing - camera alignment]{\includegraphics[width=0.4\textwidth, height=0.3\textwidth]{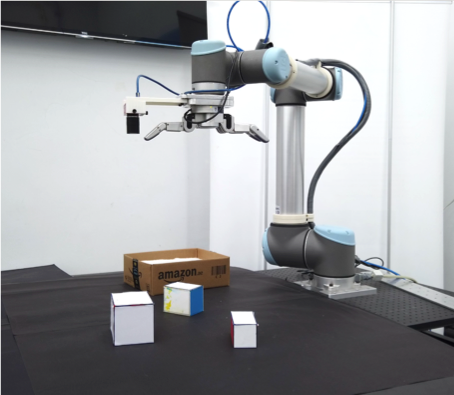}}\label{e}\hfil
\subfloat[Object and camera center matched in SAVE]{\includegraphics[width=0.4\textwidth, height=0.3\textwidth]{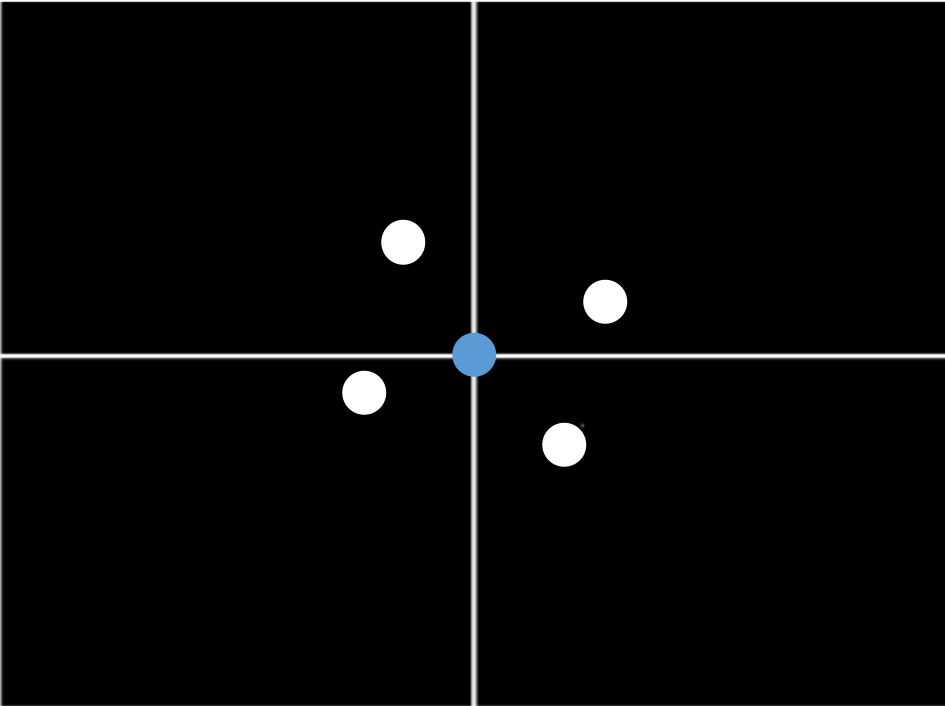}}\label{f}\\
\rotatebox{90}{Grasp, Pick and Place Stage}
\subfloat[$1^{st}$ object grasping]{\includegraphics[width=0.4\textwidth, height=0.3\textwidth]{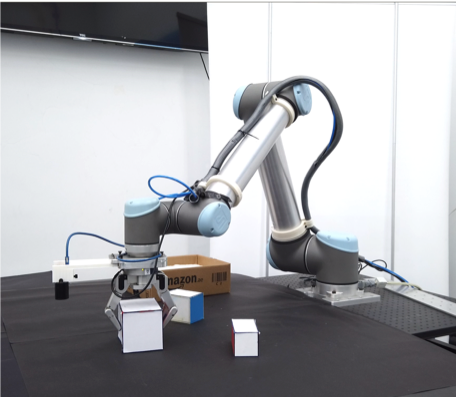}}\label{g}\hfil
\subfloat[$2^{nd}$ object grasping]{\includegraphics[width=0.4\textwidth, height=0.3\textwidth]{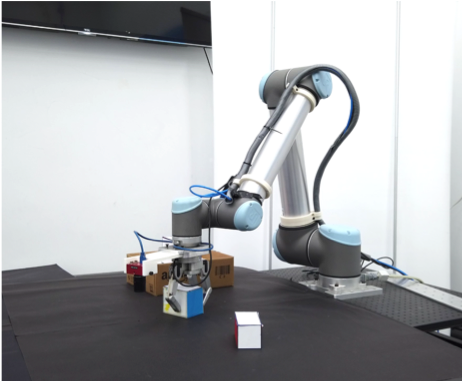}}\label{h}\\
\hfil 
\subfloat[$3^{rd}$ object grasping]{\includegraphics[width=0.4\textwidth, height=0.3\textwidth]{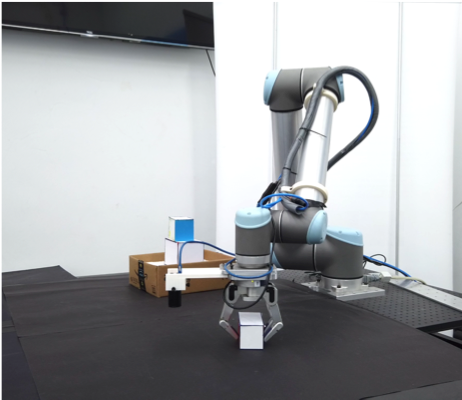}}\label{i}\hfil
\subfloat[All objects are placed into drop box]{\includegraphics[width=0.4\textwidth, height=0.3\textwidth]{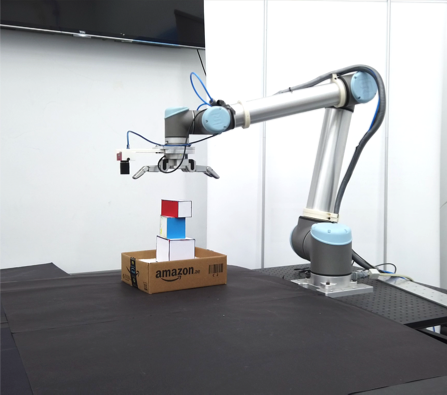}}\label{j}
\caption{Experimental sequences of proposed neuromorphic vision based multi-object grasping.}
\label{fig: experiment-size}
\end{figure*}

The five stages of the proposed neuromorphic vision-based robotic grasping approaches with multiple cubic objects of different sizes as demonstrated in Fig. \ref{fig: experiment-size}.

To quantify the grasping performance, we conducted 15 experiments of hexahedron objects with three different sizes using both model-based and model-free approaches. The size of small, medium and large object is $15*10*12cm^3$, $15*10*10cm^3$ and $10*7*8cm^3$, respectively. For individual object, five experiments are repeated and the errors are averaged. Table \ref{tab: grasping error-size} and Table \ref{tab: grasping error-size 2} show the experimental results of grasping error and object deviation of the model-free and the model-based approaches.

\begin{table}[!h]
\caption{Model-free experimental results of grasping different-size objects using event camera}
\centering 
\begin{tabular}{l|llllll}
\hline
Object Size   & $e_{gp}$ (cm) & $e_{gr}$ (degree) & $SS$ & $D_P$ (cm) & $D_R$ (degree)  & $Q_G$\\
\hline
Small         & 1.477                  & 2.14                  &  0.800   & 1.099                  & 2.10      &     0.655  \\
Medium        & 1.461                  & 2.46                  &   1.000  & 1.684                  & 1.47     &      0.530 \\
Large         & 1.498                  & 2.62                  &    1.000  & 1.343                  & 1.46      &      0.616 \\
Average/Overall & 1.479                  & 2.41                  & 0.933    &  1.375         &   1.679       &   0.600
\\
\hline
\end{tabular}
\label{tab: grasping error-size}
\end{table}
\begin{table}[!h]
\caption{Model-based experimental results of grasping different-size objects using event camera}
\centering 
\begin{tabular}{l|llllll}
\hline
Object Size   & $e_{gp}$ (cm) & $e_{gr}$ (degree) & $SS$  & $D_P$ (cm) & $D_R$ (degree)  & $Q_G$\\
\hline
Small         & 0.891                 & 4.88                  &  1.000  & 0.821 & 10.70 & 0.438   \\
Medium        & 0.742                 & 3.73                  &  1.000  & 0.361 &  0.51 &  0.893  \\
Large         & 0.481                 & 3.88                  &  1.000  & 0.711 & 2.46 &    0.740 \\
Average/Overall &   0.705               &   4.16                & 1.000   & 0.631 & 4.56 &  0.690
\\
\hline
\end{tabular}
\label{tab: grasping error-size 2}
\end{table}




Seen from Table \ref{tab: grasping error-size} and Table \ref{tab: grasping error-size 2}, 
both proposed neuromorphic vision-based grasping approaches can successfully accomplish the grasping tasks, 
and all of those evaluation metrics are within the limitations. 
By analyzing, the source of error is considered coming from several aspects.
First, the error is caused by the experimental setup that the camera is not installed exactly parallel to the work plane. 
As segmentation and tracking are executed at some height, the positioning error will occur after reaching the center at a certain height and amplified while the gripper is moving down for grasping. 
In addition, there are two manually induced errors in the grasping phase and evaluation stage. Since the object segmentation and visual servoing are accomplished in the camera frame, the position adjustment is executed after visual servoing to compensate the manually measured deviation between the centers of event camera and Barrett hand. 
The similar deviation exists in the evaluation system, while calculating the optimal grasping pose by transformation from ArUco marker center to object top surface center.  
Besides, the low spatial resolution of DAVIS 346C utilized can also cause the error. 

\subsubsection{Robustness Testing}
To test the robustness of the proposed grasping approaches using an event camera, the additional experiments were conducted in low-light condition and using objects of other shapes. 

\paragraph{\textbf{low-light conditions}}
One of the advantages of an event camera is high sensitivity to the change of light intensity, that can observe objects even in the low-light environment. However, more noise will also be captured in low-light condition. So the noise filter is applied to eliminate the noise and capture more meaningful events. We conducted 5 experiments for each cubic/hexahedron object. 
The experimental results of model-free and model-based approaches are recorded in Table \ref{tab: grasping error-size}, including grasping pose error in terms of $e_{gp}$ and $e_{gr}$, and the object deviation in terms of $D_P$ and $D_R$. 
The success rate $SS$ and grasp quality $Q_G$ are also calculated with the same position limitation $L_P=2cm$ and orientation limitation $L_R=15^{\circ}$. 

\begin{table}[!h]
\caption{Model-free experimental results of grasping different-size objects using event camera in low-light environment}
\centering 
\begin{tabular}{l|llllll}
\hline
Object Size   & $e_{gp}$ (cm) & $e_{gr}$ (degree) & $SS$ & $D_P$(cm) & $D_R$ (degree) & $Q_G$  \\
\hline
Small         & 1.443                  & 2.61       &     1.000    &   1.373                &      2.74     &    0.565         \\
Medium        & 1.551                  & 2.91       &      1.000    &    1.046              &      2.32      &   0.661     \\
Large         & 1.411                  & 2.88       &       0.600     &    1.203              &      1.96      &   0.503   \\
Average/Overall &  1.478             & 2.80       &     0.867  &  1.207               &      2.34 &    0.576 
\\
\hline
\end{tabular}
\label{tab: grasping error-size}
\end{table}
\begin{table}[!h]
\caption{Model-based experimental results of grasping different-size objects using event camera in low-light environment}
\centering 
\begin{tabular}{l|llllll}
\hline
Object Size   & $e_{gp}$ (cm) & $e_{gr}$ (degree) & $SS$ & $D_P$(cm) & $D_R$ (degree) & $Q_G$\\
\hline
Medium         &    0.951            &    5.41    & 1.000   & 0.40 & 3.121  &  0.795      \\
Large        &    1.120             &      6.22    & 1.000   &  1.05 & 5.151  &   0.566     \\
Average/Overall &   1.040             &     5.82   & 1.000   & 0.73 & 4.135  & 0.681       
\\
\hline
\end{tabular}
\label{tab: grasping error-size 2}
\end{table}

\begin{figure*}[]
    \hspace*{-0.3cm} 
    \begin{subfigure}[t]{0.5\textwidth}
        \centering
        \includegraphics[height=1.9in]{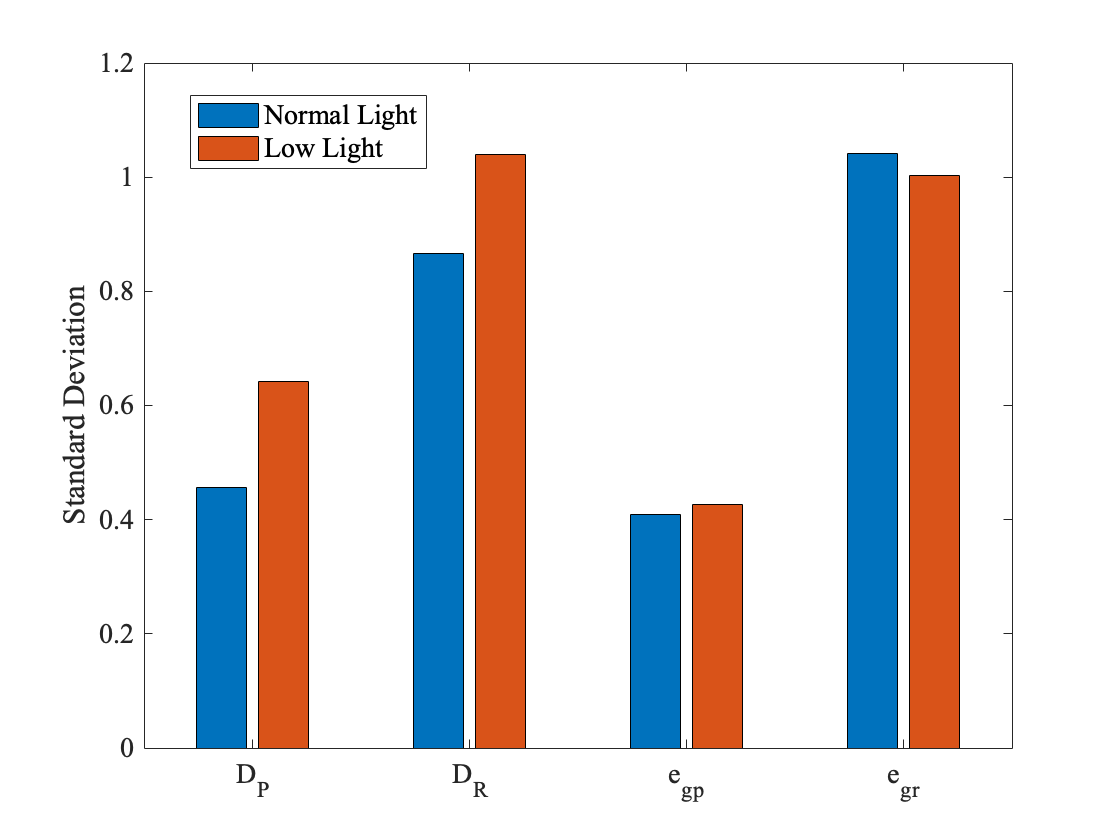}
        \caption{Model-free approach}
    \end{subfigure}%
    ~ 
    \begin{subfigure}[t]{0.4\textwidth}
        \centering
        \includegraphics[height=1.9in]{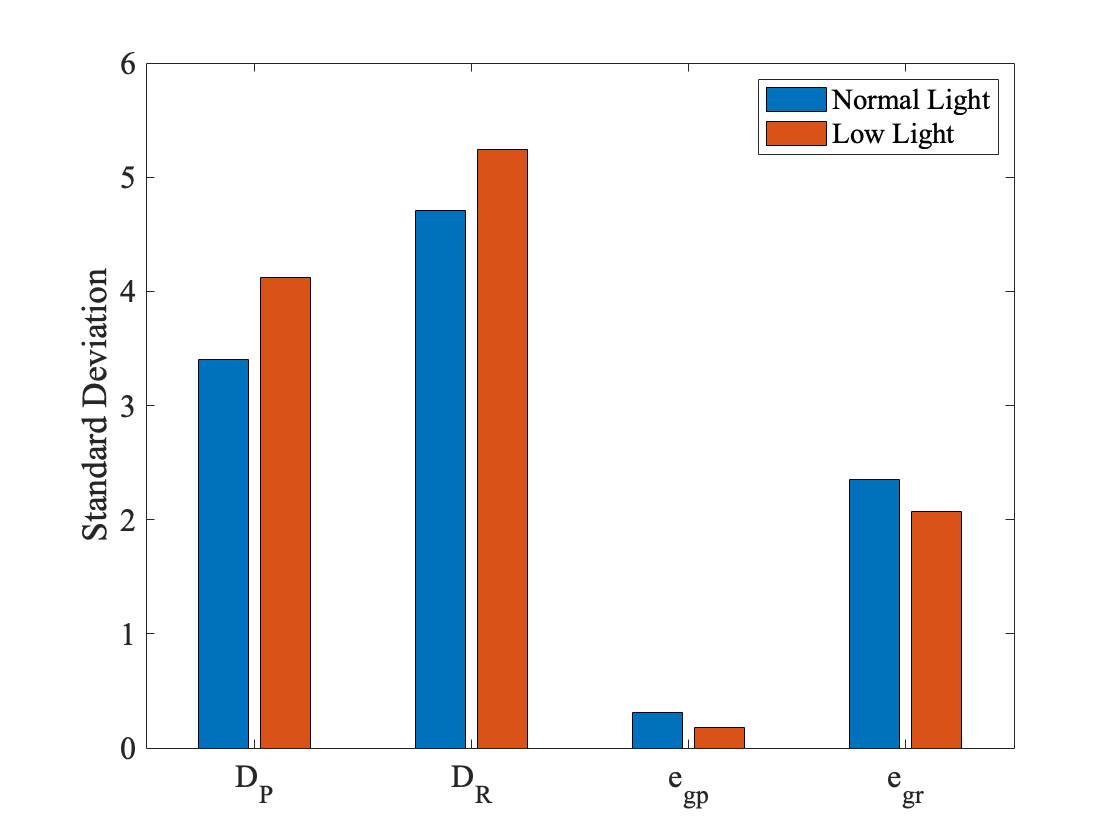}
        \caption{Model-based approach}
    \end{subfigure}
    \caption{Standard deviation of grasping errors and object deviations in normal-light and low-light condition by model-free approach}
    \label{fig: SD}
\end{figure*}

Seen from the results, the average errors are all within the limitations and the success rate and grasp quality score are similar to those in the normal light environment. However, by comparing the standard deviation of the model-free approach as shown in Fig. \ref{fig: SD}(a), the overall performance in the low-light condition is more unstable with a higher standard deviation, even though the value of position error of grasping orientation under normal light is slightly higher. For the model-based approach, it presents a higher standard deviation value of object deviation in low-light condition as depicted in Fig. \ref{fig: SD}(b). On the whole, both of our proposed approaches can reach the grasping goal successfully in low-light environment. 

\begin{figure*}[!h]
    \hspace*{-0.3cm} 
    \begin{subfigure}[t]{0.5\textwidth}
        \centering
        \includegraphics[height=1.9in]{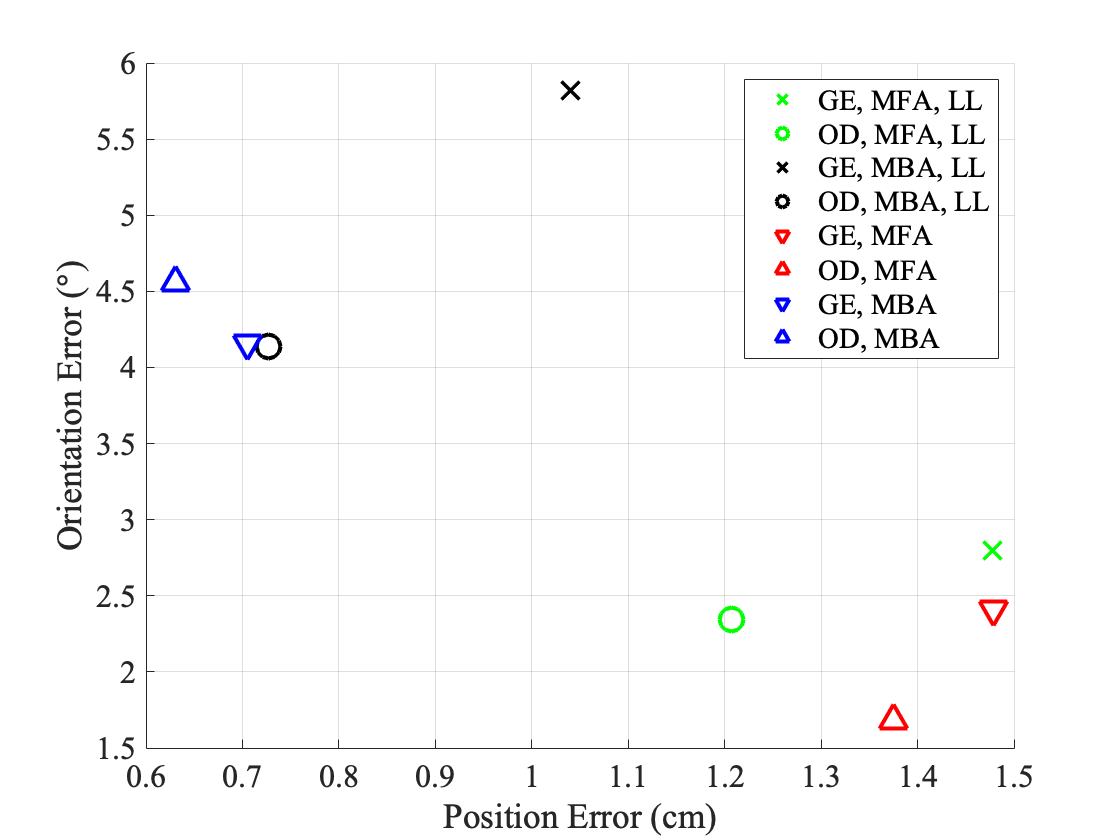}
        \caption{Grasping error and object deviation}
    \end{subfigure}%
    ~ 
    \begin{subfigure}[t]{0.4\textwidth}
        \centering
        \includegraphics[height=1.9in]{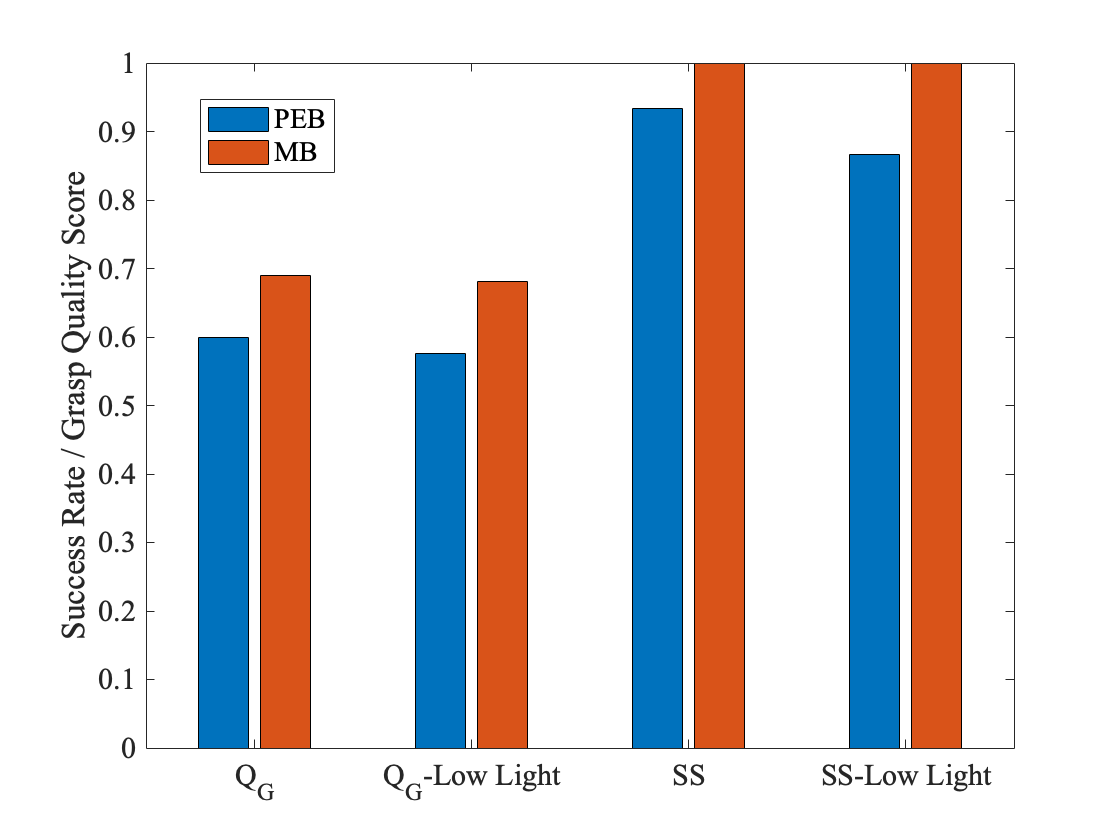}
        \caption{Success rate and grasp quality score}
    \end{subfigure}
    \caption{Comparison of the proposed model-based and model-free approaches}
    \label{fig: model-event}
\end{figure*}

The comparison between the two proposed approaches is depicted in Fig. \ref{fig: model-event}, where MFA and MBA presents the model-free approach and model-based approach, GE and OD indicates the grasping error and object deviation, and LL expresses the low-light condition. From Fig. \ref{fig: model-event} (a), the model-free approach reaches a relative smaller orientation error and a larger position error comparing to the model-based approach. In both low-light and normal-light conditions, the model-based approaches reaches a higher successful rate and grasp quality score as indicated in Fig. \ref{fig: model-event} (b). 


\begin{figure*}[!h]
    \centering
    \begin{subfigure}[t]{0.5\textwidth}
        \centering
        \includegraphics[height=1.4in]{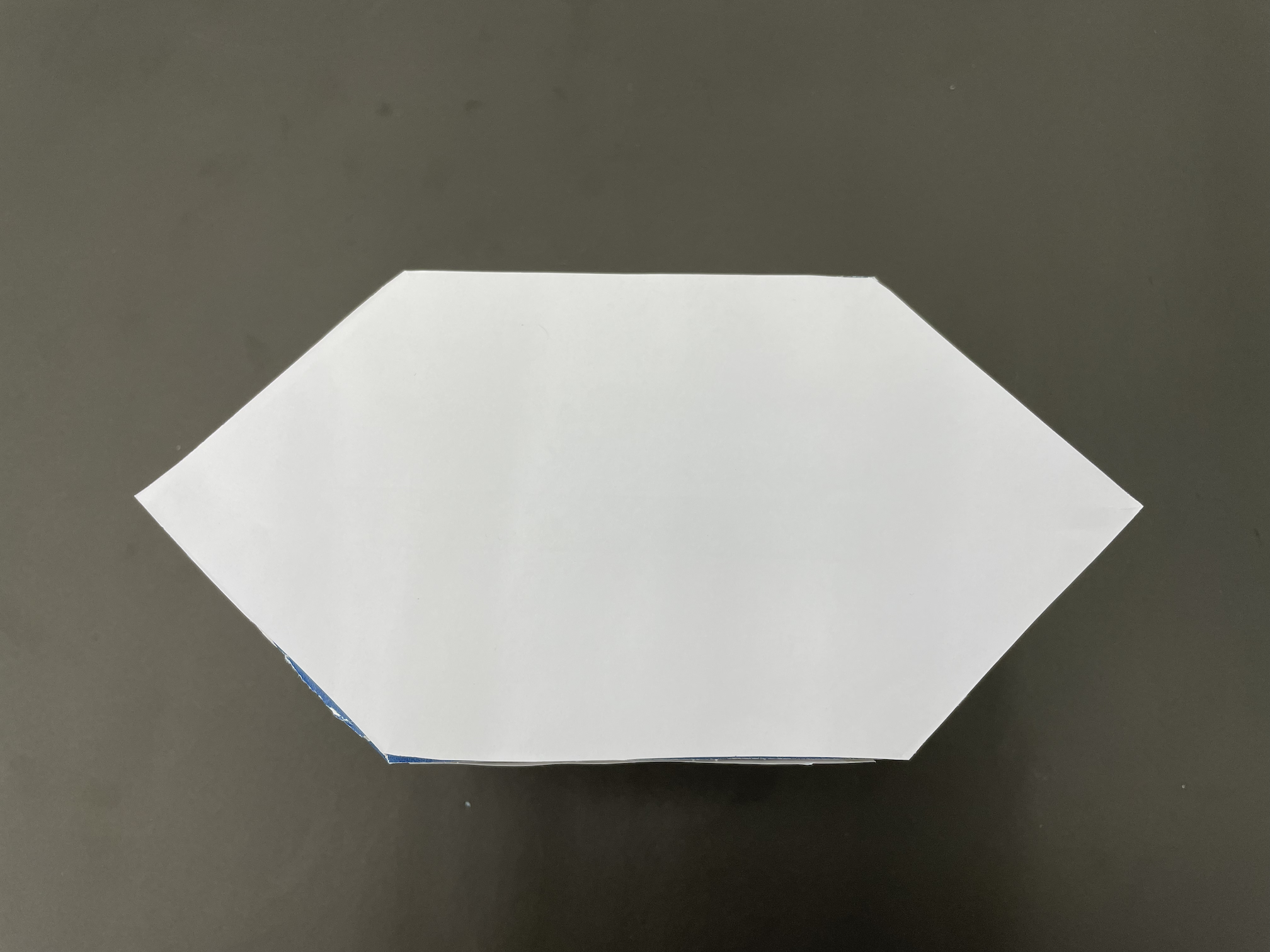}
        \caption{Octahedron-6: top view}
    \end{subfigure}%
    ~ 
    \begin{subfigure}[t]{0.5\textwidth}
        \centering
        \includegraphics[height=1.4in]{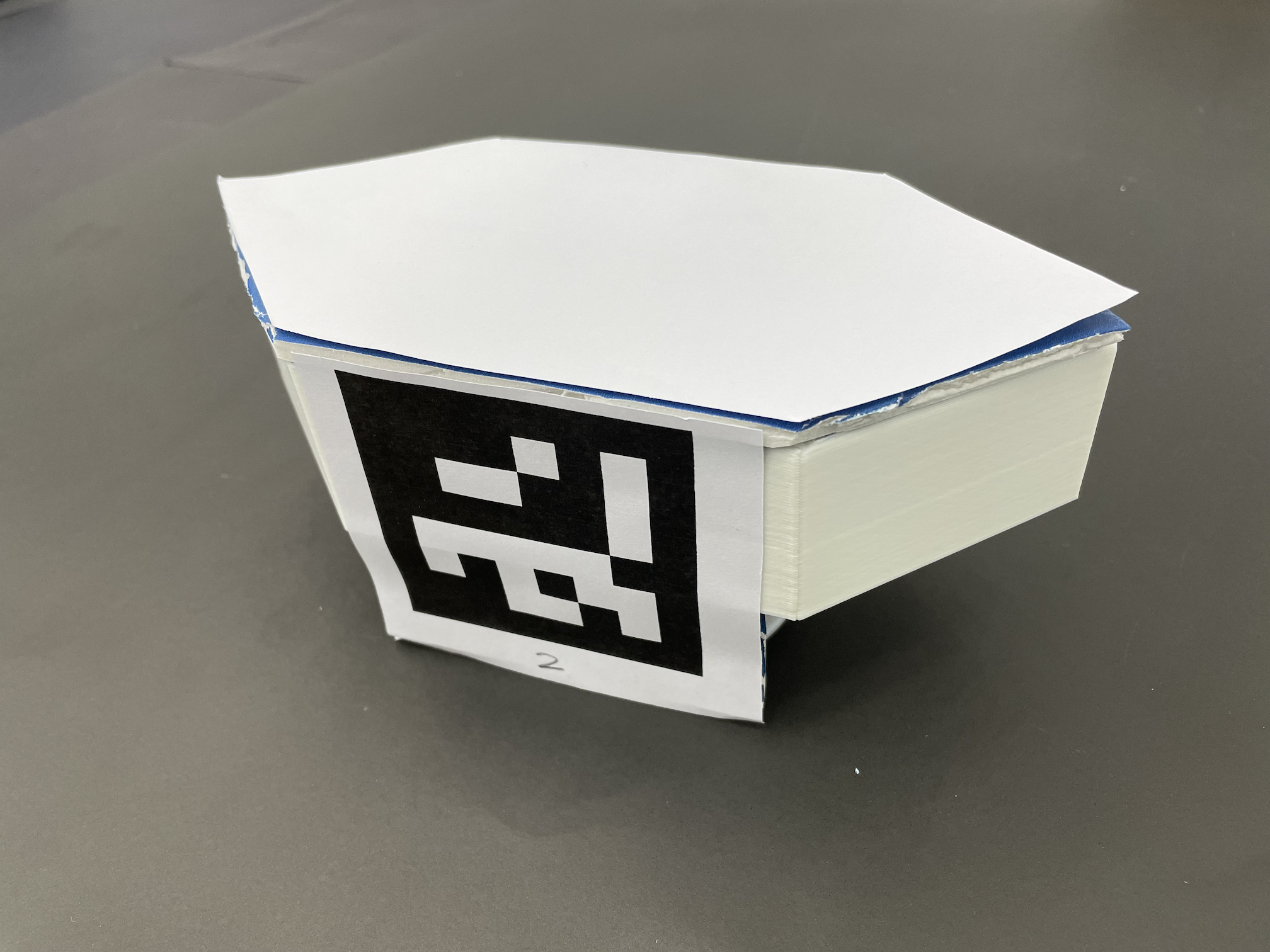}
        \caption{Octahedron-6: side view}
    \end{subfigure}
    ~
    \begin{subfigure}[t]{0.5\textwidth}
        \centering
        \includegraphics[height=1.4in]{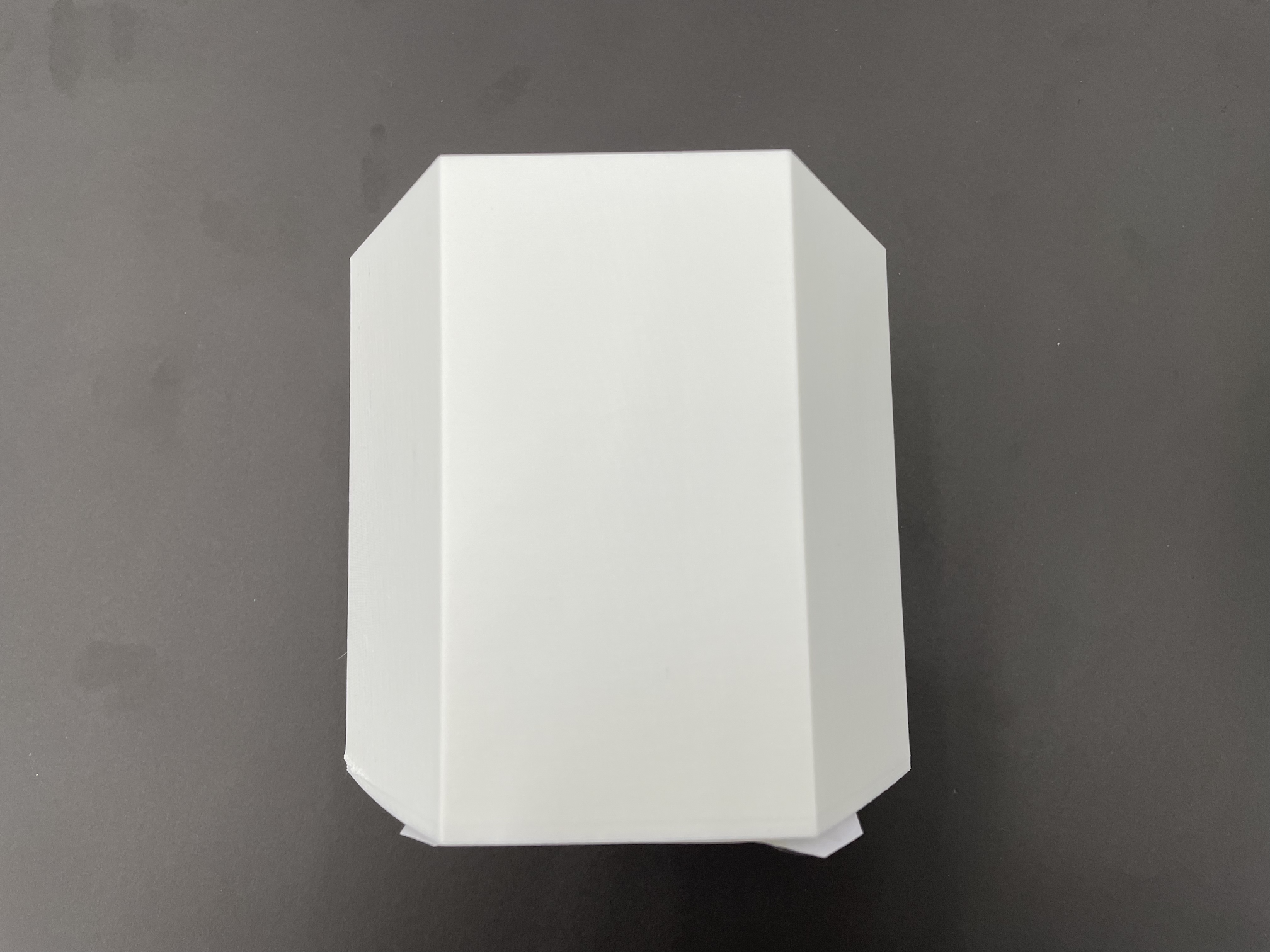}
        \caption{Octahedron-8: top view}
    \end{subfigure}%
    ~ 
    \begin{subfigure}[t]{0.5\textwidth}
        \centering
        \includegraphics[height=1.4in]{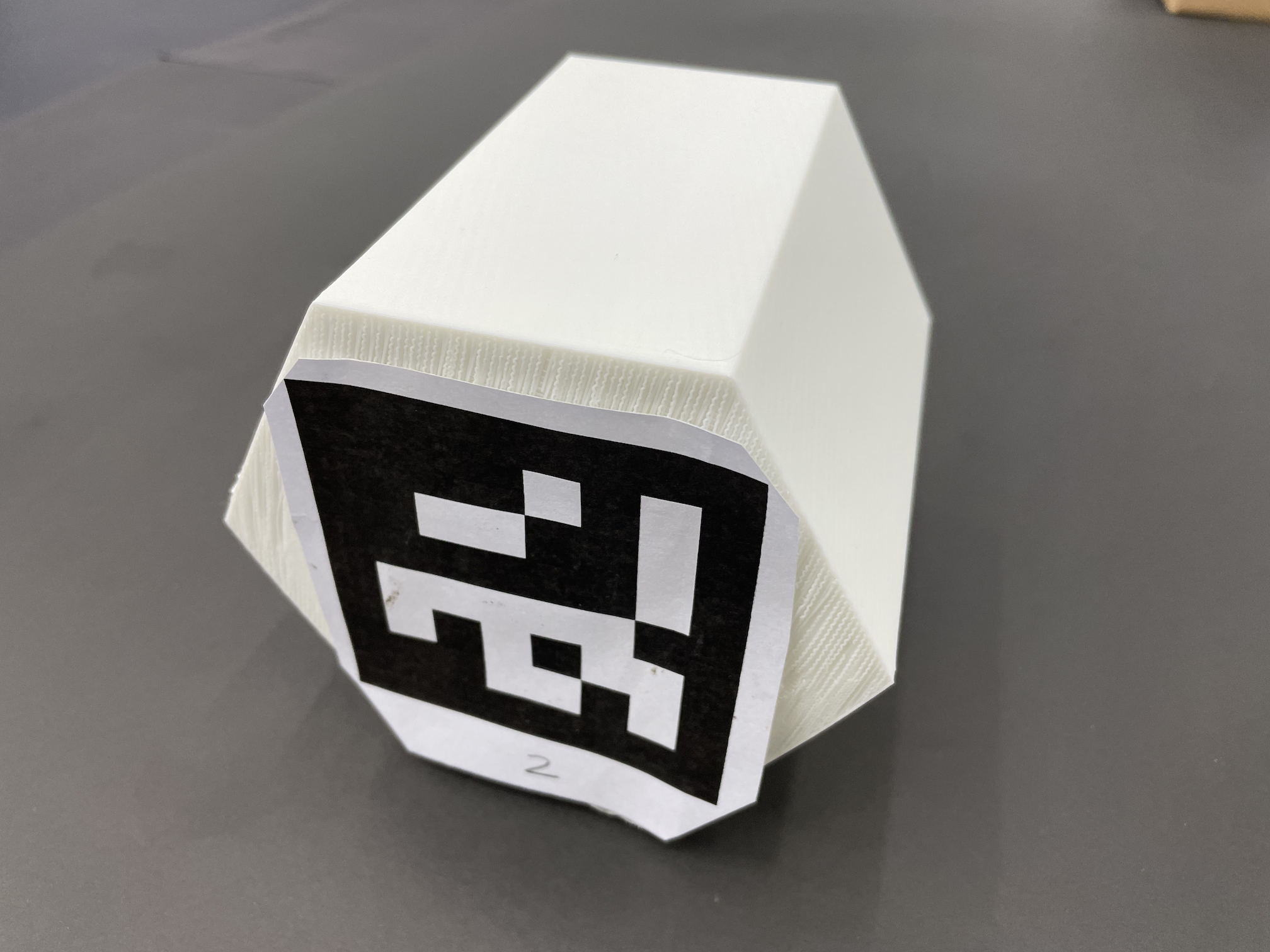}
        \caption{Octahedron-8: side view}
    \end{subfigure}
    \caption{Two octahedrons utilized in grasping experiments. Octahedrons with six corners (Octahedron-6) in the top view (a) and in the side view (b). Octahedrons with eight corners (Octahedron-8) in the top view (c) and in the side view (d). }
    \label{fig: octahedron}
\end{figure*}

\paragraph{\textbf{Objects with different shapes}}
In addition, the experiments of grasping different shapes were also conducted to test the robustness. Besides of hexahedron, two octahedrons with different shapes demonstrated in Fig. \ref{fig: octahedron} are utilized as unknown objects to validate the robustness of the proposed approach. 

Since the model-based approach can only be applied to known objects with the prior model, only the model-free approach can achieve this task as the two octahedrons are considered as unknown objects. Table \ref{tab: grasping shape} shows the grasping pose error, object deviation, success rate and grasp quality of experiments on objects with varying shapes. 
Octahedron-6 and octahedron-8 represent the octahedron with six and eight visible corner features from the top view as shown in Fig. \ref{fig: octahedron}.

\begin{table}[!h]
\caption{Experimental results of grasping different-shape objects by model-free approach using event camera}
\centering 
\begin{tabular}{l|llllll}
\hline
Object Shape   & $e_{gp}$ (cm) & $e_{gr}$ (degree) & $SS$ & $D_P$ (cm) & $D_R$ (degree) & $Q_G$\\
\hline
heptahedron          &    1.479                   &  2.408      &     0.933     &   1.375              &   1.679     &  0.600\\
octahedron-6         &    1.557            &    2.816           &     1.000  &   0.901         &   2.040        & 0.707  \\
octahedron-8         &   1.530                 &   2.793        &      0.800     &   0.993       &   2.431      &   0.671  \\
Average/Overall &   1.522 &    2.672         &  0.880    &    1.090     &  2.05  &   0.659  
\\
\hline
\end{tabular}
\label{tab: grasping shape}
\end{table}

\begin{figure*}[!]
    \centering
    \begin{subfigure}[t]{0.8\textwidth}
        \centering
        \includegraphics[height=1.8in]{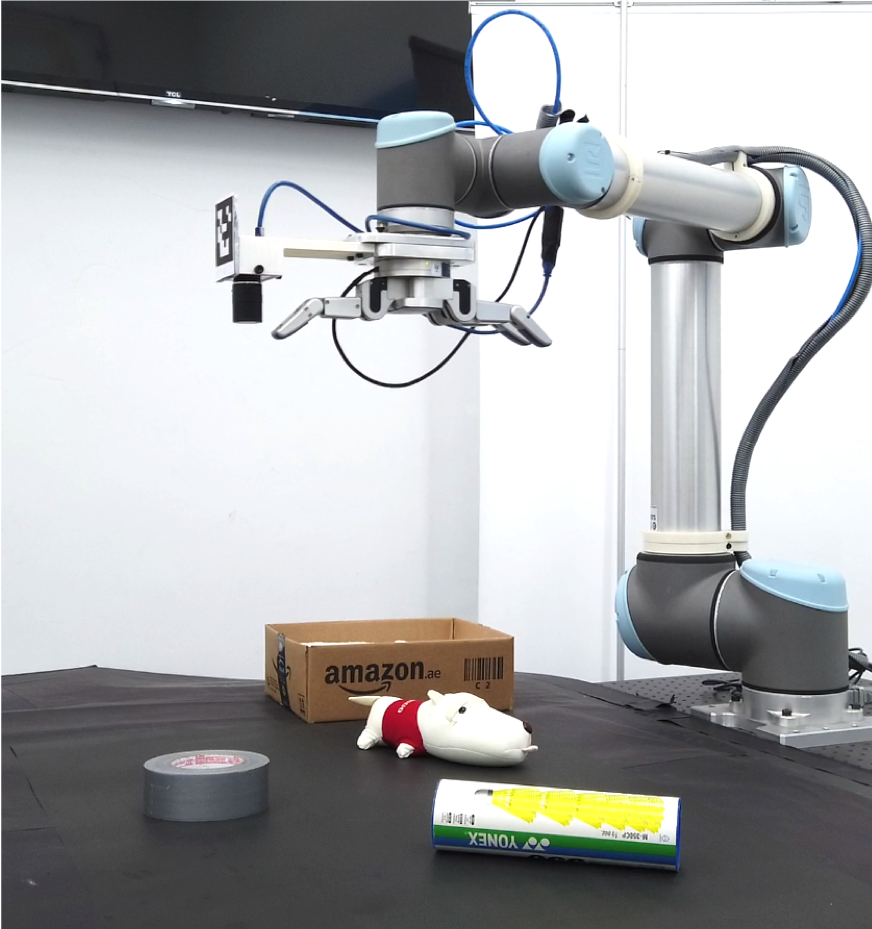}
        \caption{initial state with real objects: soft doll, badminton tube and tape}
    \end{subfigure}%
    \hfill
    ~ 
    \begin{subfigure}[t]{0.4\textwidth}
        \centering
        \includegraphics[height=1.8in]{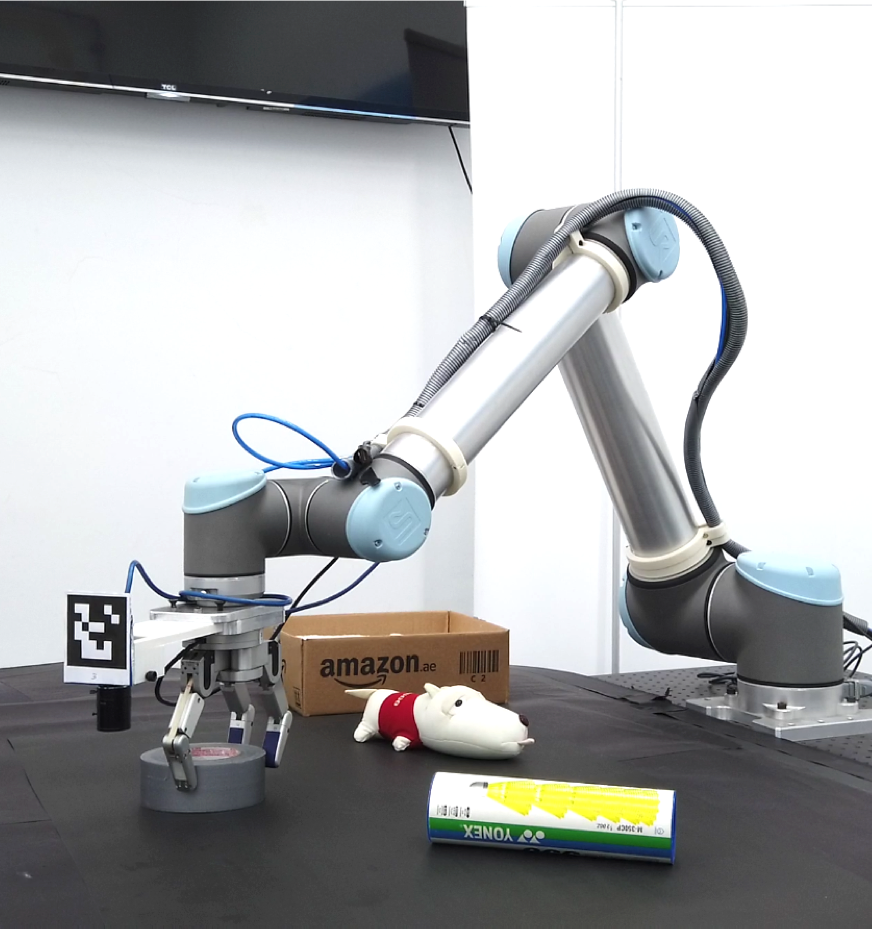}
        \caption{grasp tape}
    \end{subfigure}%
    ~ 
    \begin{subfigure}[t]{0.4\textwidth}
        \centering
        \includegraphics[height=1.8in]{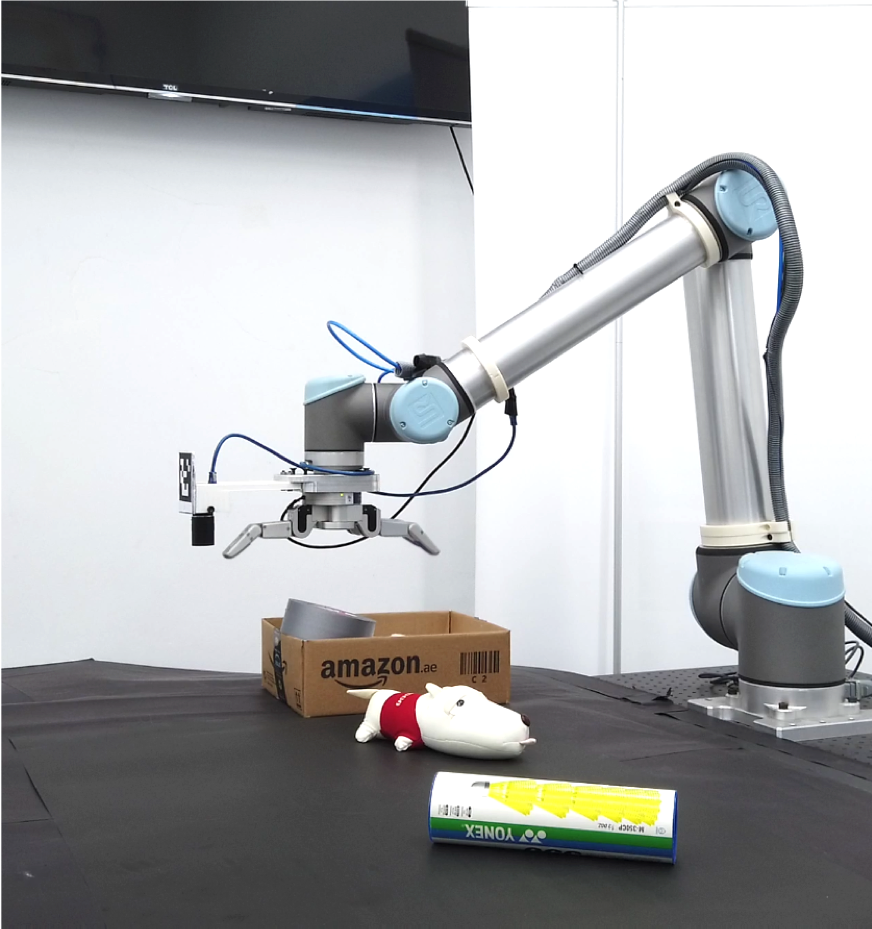}
        \caption{drop tape}
    \end{subfigure}
    \hfill
    ~
    \begin{subfigure}[t]{0.4\textwidth}
        \centering
        \includegraphics[height=1.8in]{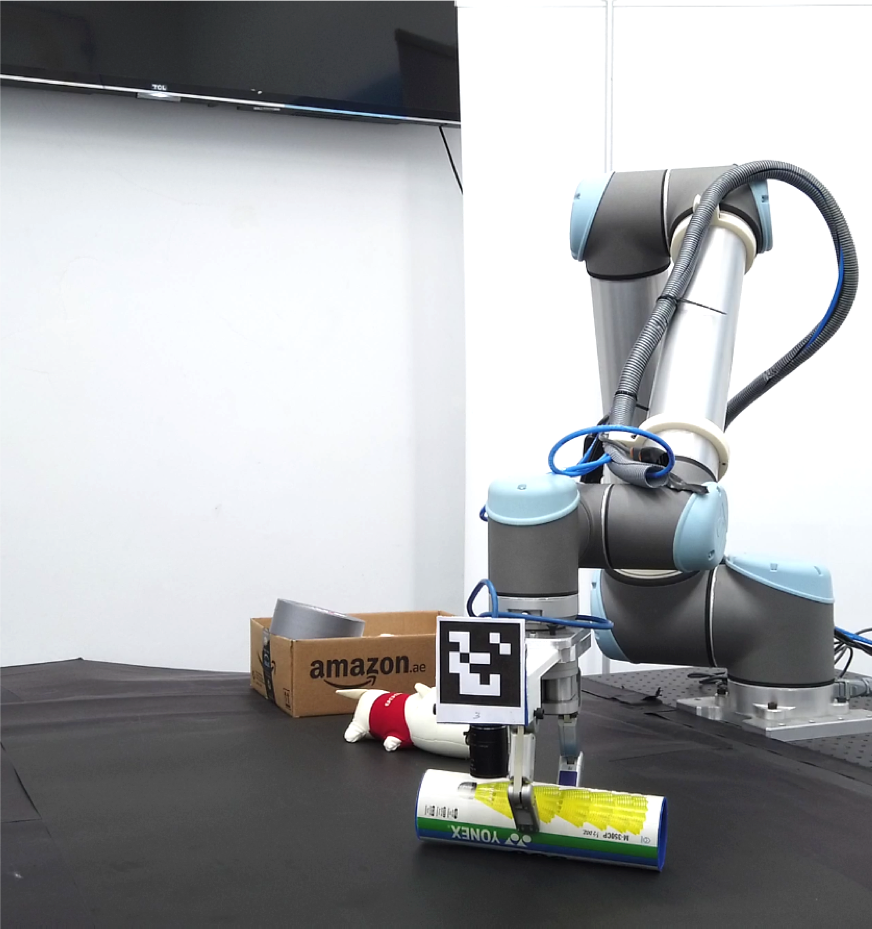}
        \caption{grasp badminton tube}
    \end{subfigure}%
    ~
    \begin{subfigure}[t]{0.4\textwidth}
        \centering
        \includegraphics[height=1.8in]{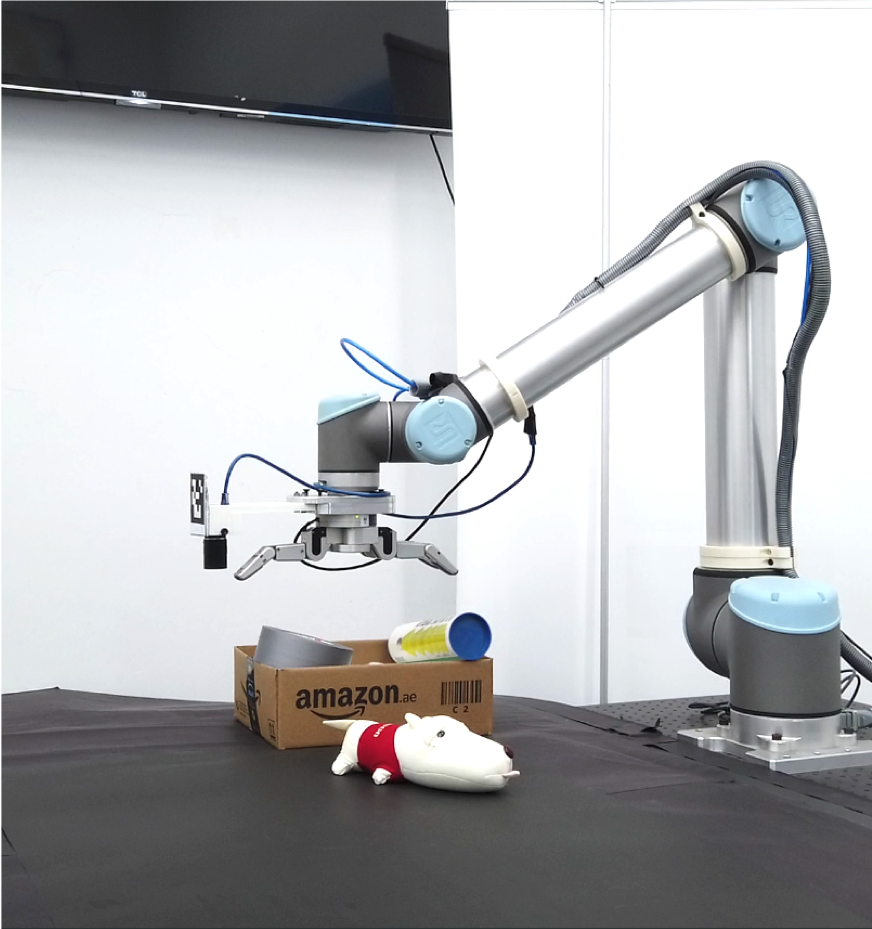}
        \caption{drop badminton tube}
    \end{subfigure}%
    \hfill
    ~
    \begin{subfigure}[t]{0.4\textwidth}
        \centering
        \includegraphics[height=1.8in]{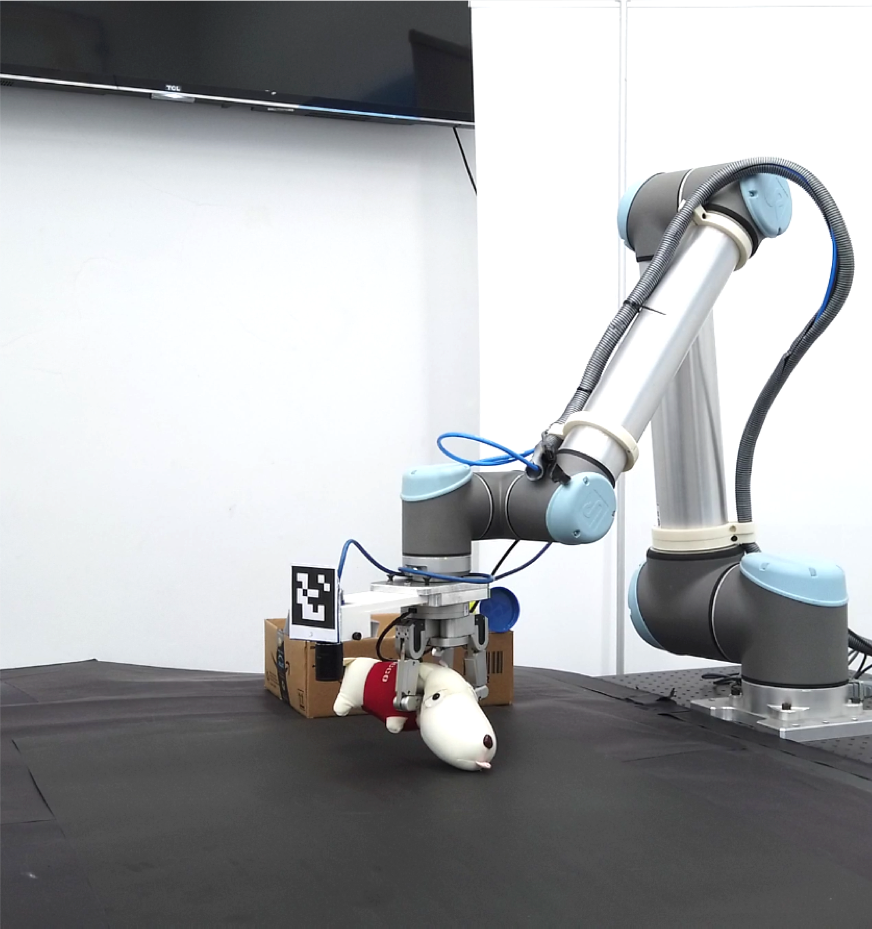}
        \caption{grasp soft doll}
    \end{subfigure}%
    ~
    \begin{subfigure}[t]{0.4\textwidth}
        \centering
        \includegraphics[height=1.8in]{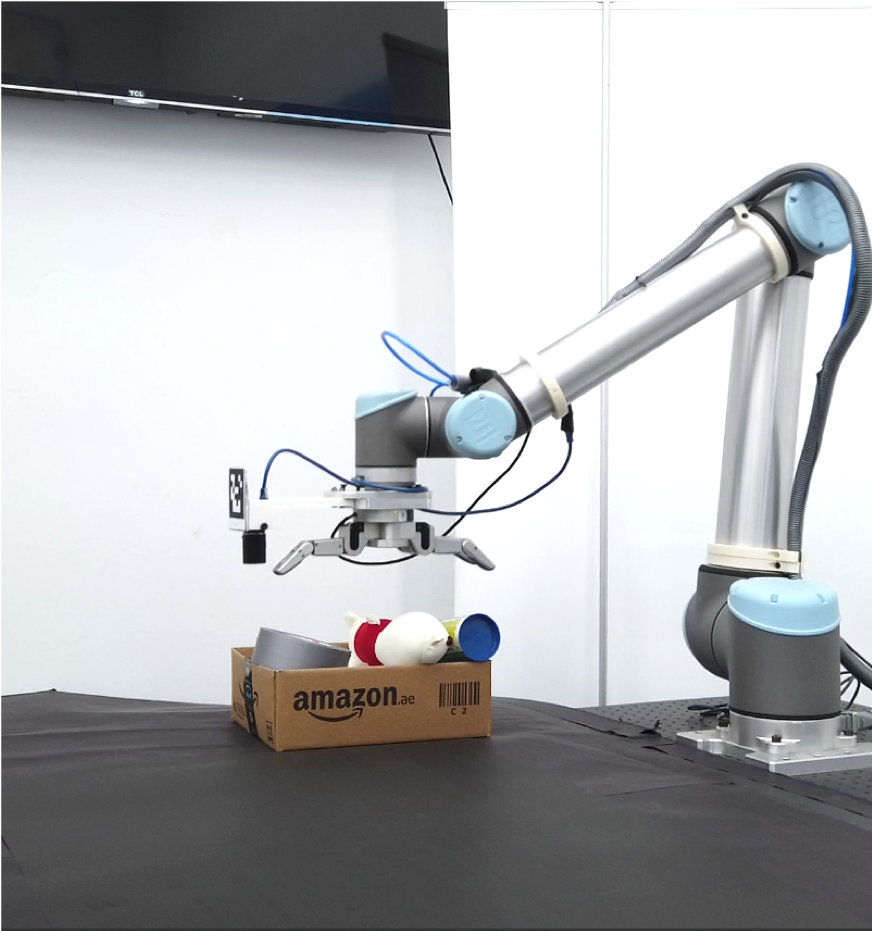}
        \caption{drop soft doll}
    \end{subfigure}
    \caption{Picking and placing process of real objects by the proposed neuromorphic vision based multi-object grasping approach.}
    \label{fig: real objects}
\end{figure*}

By experimental validation, all those objects with different shapes can be pick and placed effectively. Seen from Table \ref{tab: grasping shape}, 
the grasping of three objects of different shapes demonstrates a comparable performance. Furthermore, the proposed model-free approach is also validated on real objects in daily life. 
The whole pick and place process is demonstrated in Fig. \ref{fig: real objects}, that achieves a successful pick and place task for multiple objects.

\section{Discussion}
\label{sec: discussion}


Both model-based and model-free approaches proposed are valid for multiple-object grasping using an event camera. By comparison, the pros and cons of the two approaches are concluded in Table \ref{tab: summary}.
\begin{table}[!h]
\caption{Comparison of the proposed model-based and model-free approaches}
\begin{tabular}{l|ll}
\hline
Terms & Model-based Approach & Model-free Approach \\ \hline
Pros  &    Higher accuracy   &      \begin{tabular}[c]{@{}l@{}}Model free\\ Unknown and moving objects \\ Robust to imperfect perception\end{tabular}                      \\
\hline
Cons  &   \begin{tabular}[c]{@{}l@{}}Prior knowledge of model is required\\ Sensitive to imperfect perception\end{tabular}    &         Relatively lower accuracy                    \\ \hline
\end{tabular}
\label{tab: summary}
\end{table}

From the experimental results in Section \ref{sec: results}, the model-based approach shows a slightly better grasping performance with less error because of the position based visual servoing. But the model-based approach is limited to known objects as it requires offline modeling of objects. However, the objects are generally unknown requiring online process in real scenarios. It indicates that the model-based approach more suitable for grasping tasks for the specific or pre-defined objects. 
By contrast, the proposed model-free approach can obtain the position information of unknown objects without prior knowledge, which shows a great advantage in practical and real applications. Moreover, it is quite less sensitive to the imperfect perception than the model-based approach. In addition, the model-free approach also shows the flexibility and possibility to deal with the moving object besides of the static object.

\section{Conclusion}
We proposed an event-based grasping framework for robotic manipulator with neuromorpihc eye-in-hand configuration. Particularly, a model-based and a model-free approaches for multiple-object grasping in a cluttered scene are developed. The model-based approach provides a solution for grasping known objects in the environment, with prior knowledge of the object shape to be grasped. It consists of the 3D reconstruction of the scene, euclidean distance clustering, position-based visual servoing and grasp planning. 
Differently, the model-free approach can be applied to unknown objects grasping applications in real time, which consists of the developed event-based segmentation, visual servoing adopting depth information and grasp plan. 

By experimentally validating with objects of different sizes and in different light conditions, both approaches can effectively achieve the multiple-object grasping task successfully. From the quantity evaluation of the grasping pose, success rate, object deviation and grasp quality, the model-based approach presents slightly more accurate because of the position-based visual servoing. 
However, the model-based approach is constrained to known objects with prior knowledge of models. The model-free approach is more applicable in real scenarios for operating unknown objects, which is experimentally validated with real objects in this paper. To conclude, both model-based and model-free are applicable and effective for neuromorphic vision-based multiple-object grasping applications, which can boost production speed in factory automation. According to their pros and cons, the particular approach can be selected in different specific scenario. This paper demonstrates grasping for multi-object in simple scenarios, we will focus on the event-based object segmentation for more complex situations such as objects with occlusion in the future. 

\begin{acknowledgements}
This work is supported by the Khalifa University of Science and Technology under Award No. CIRA-2018-55 and RC1- 2018-KUCARS, and was performed as part of the Aerospace Research and Innovation Center (ARIC), which is jointly funded by STRATA Manufacturing PJSC (a Mubadala company) and Khalifa University of Science and Technology
\end{acknowledgements}

%


%
%


%
%

\bibliographystyle{unsrt}
\bibliography{references}


\end{document}